\documentclass[conference]{IEEEtran}
\IEEEoverridecommandlockouts
\usepackage{times}
\usepackage{soul}
\usepackage{url}
\usepackage[hidelinks]{hyperref}
\usepackage[utf8]{inputenc}
\usepackage[small]{caption}
\usepackage{graphicx}
\usepackage{amsthm}
\usepackage{amsmath}
\usepackage{amssymb}
\usepackage{booktabs}
\usepackage{algorithm}
\usepackage{algorithmicx,algpseudocode}

\let\oldReturn\Return
\renewcommand{\Return}{\State\oldReturn}

\usepackage[super]{nth}
\usepackage{bm}
\usepackage{subcaption}
\usepackage{mathtools}
\usepackage{tcolorbox}
\usepackage{enumitem}
\usepackage{multirow}
\usepackage{anyfontsize}
\usepackage{adjustbox}

\def\eqref#1{equation~\ref{#1}}
\def\1{\bm{1}}

\DeclareMathAlphabet{\mathsfit}{\encodingdefault}{\sfdefault}{m}{sl}
\SetMathAlphabet{\mathsfit}{bold}{\encodingdefault}{\sfdefault}{bx}{n}

\allowdisplaybreaks

\DeclareMathOperator*{\argmax}{argmax}
\DeclareMathOperator*{\argmin}{argmin}

\newtheorem{theorem}{Theorem}

\newtheorem{definition}{Definition}

\def\BibTeX{{\rm B\kern-.05em{\sc i\kern-.025em b}\kern-.08em
    T\kern-.1667em\lower.7ex\hbox{E}\kern-.125emX}}
\begin{document}

\setlength{\textfloatsep}{5pt}
\setlength{\dbltextfloatsep}{5pt}

\title{Block-Structured Optimization for Subgraph Detection in Interdependent Networks
% {\footnotesize \textsuperscript{*}Note: Sub-titles are not captured in Xplore and
% should not be used}
% \thanks{Identify applicable funding agency here. If none, delete this.}
}

\author{
\IEEEauthorblockN{Fei Jie\IEEEauthorrefmark{1}\IEEEauthorrefmark{2}\thanks{The first two authors contributed equally to this research.}, Chunpai Wang\IEEEauthorrefmark{3}, Feng Chen\IEEEauthorrefmark{4}, Lei Li\IEEEauthorrefmark{1}\IEEEauthorrefmark{2}, Xindong Wu\IEEEauthorrefmark{5}\IEEEauthorrefmark{1}\IEEEauthorrefmark{6}}
\IEEEauthorblockA{\IEEEauthorrefmark{1}
Key Laboratory of Knowledge Engineering with Big Data (Hefei University of Technology), Ministry of Education, Hefei, China.}
\IEEEauthorblockA{\IEEEauthorrefmark{2}School of Computer Science and Information Engineering, Hefei University of Technology, Hefei, China.}
\IEEEauthorblockA{\IEEEauthorrefmark{3}Department of Computer Science, University at Albany – SUNY, Albany, NY, USA}
\IEEEauthorblockA{\IEEEauthorrefmark{4}Erik Jonsson School of Engineering \& Computer Science, 
The University of Texas at Dallas, Dallas, TX, USA}
\IEEEauthorblockA{\IEEEauthorrefmark{5}Mininglamp Academy of Sciences, Mininglamp Technologies, Beijing, China}
\IEEEauthorblockA{\IEEEauthorrefmark{6}Institute of Big Knowledge Science, Hefei University of Technology, Hefei, China}
realfjie@gmail.com,
cwang25@albany.edu,
feng.chen@utdallas.edu, %fchen5@albany.edu
\{lilei,xwu\}@hfut.edu.cn
% \IEEEauthorblockN{Fei Jie}
% \IEEEauthorblockA{\textit{Department of Computer Science} \\
% \textit{Hefei University of Technology}\\
% Hefei, China \\
% hfut\_jf@aliyun.com}
% \and
% \IEEEauthorblockN{Chunpai Wang}
% \IEEEauthorblockA{\textit{Department of Computer Science} \\
% \textit{University at Albany – SUNY}\\
% Albany, USA \\
% cwang25@albany.edu}
% \and
% \IEEEauthorblockN{Feng Chen}
% \IEEEauthorblockA{\textit{Department of Computer Science} \\
% \textit{University at Albany – SUNY}\\
% Albany, USA \\
% fchen5@albany.edu}
% \and
% \IEEEauthorblockN{Lei Li}
% \IEEEauthorblockA{\textit{Department of Computer Science} \\
% \textit{Hefei University of Technology}\\
% City, Country \\
% lilei@hfut.edu.cn}
% \and
% \IEEEauthorblockN{Xindong Wu}
% \IEEEauthorblockA{\textit{Department of Computer Science} \\
% \textit{Hefei University of Technology}\\
% City, Country \\
% xwu@hfut.edu.cn}
% \IEEEauthorblockN{Anonymous Submission}
% \IEEEauthorblockA{\textit{Anonymous affiliations} \\
% Anonymous emails}
}
\maketitle

\begin{abstract}

% \textcolor{red}{Many different physical, biological and social systems are structured as networks. Often, these networks interact with others in non-trivial ways and form interdependent networks. 
% % \textcolor{red}{It has been assumed that functional failure of nodes in one network may result in failure of dependent nodes belonging to different networks. Detecting subgraphs related to those failure nodes or significant nodes in such networks could reveal hidden patterns, gain deep insights of  the complex system, and prevent potential functional failure in the near future. Subgraph detection in interdependent network is challenging due to noised signals and large scale networks.}
% It is important to detect significant or abnormal subgraphs across multiple networks which are interdependent to gain deep insights of the complex system. In this paper,} 

We propose a generalized framework for block-structured nonconvex optimization, which can be applied to structured subgraph detection in interdependent networks, such as multi-layer networks, temporal networks, networks of networks, and many others. Specifically, we design an effective, efficient, and parallelizable projection algorithm, namely Graph Block-structured Gradient Projection (GBGP), to optimize a general non-linear function subject to graph-structured constraints. We prove that our algorithm: 1) runs in nearly-linear time on the network size; 2) enjoys a theoretical approximation guarantee. Moreover, we demonstrate how our framework can be applied to two very practical applications
% including anomalous evolving subgraph detection and subgraph detection in network of networks, 
and conduct comprehensive experiments to show the effectiveness and efficiency of our proposed algorithm.
\end{abstract}

\begin{IEEEkeywords}
subgraph detection, sparse optimization, interdependent networks
\end{IEEEkeywords}

\section{Introduction}
\begin{figure}
    \centering
    \begin{subfigure}[t]{0.20\textwidth}
        \centering
        \includegraphics[width=\linewidth,height=0.85\linewidth]{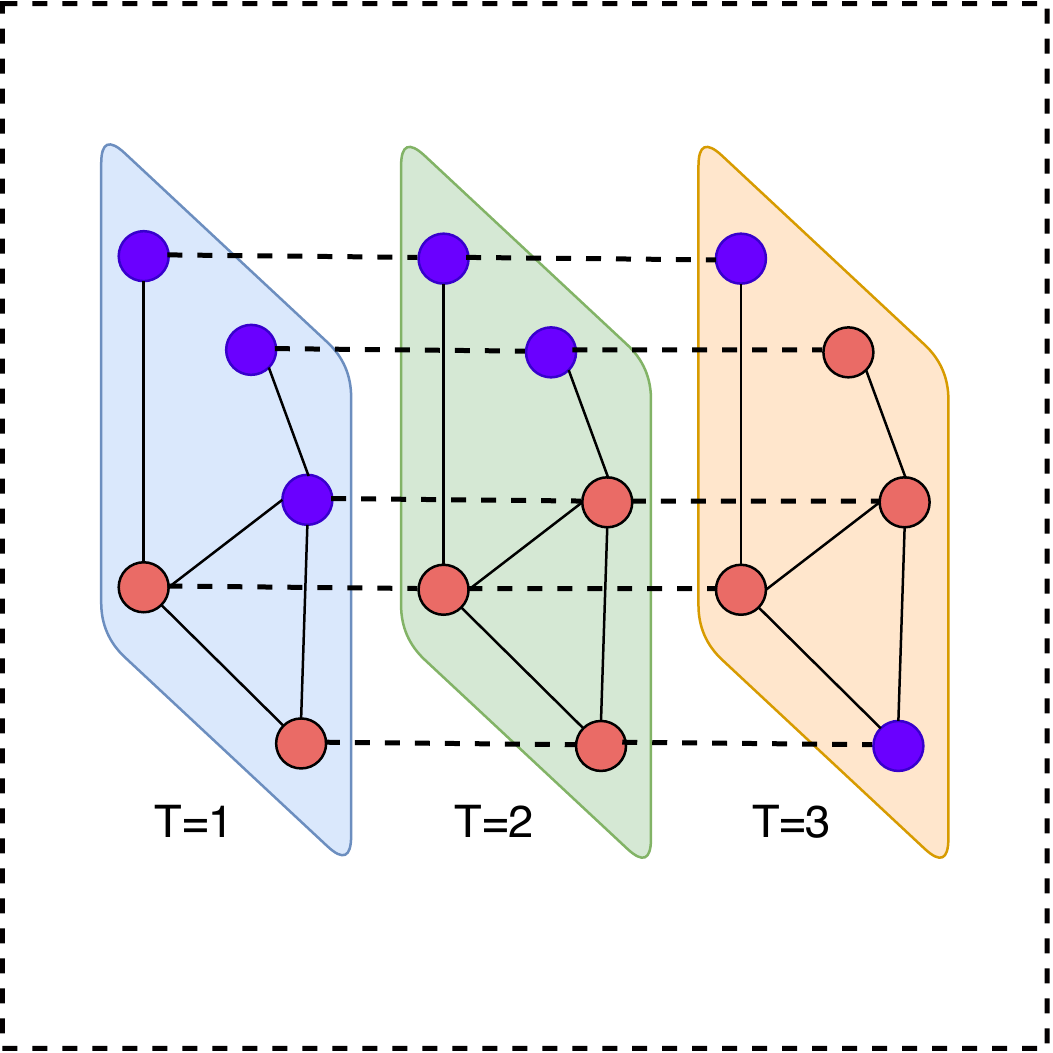}
        \caption{Temporal Networks}
       \label{fig:temporal_network}
    \end{subfigure}
    \hspace{23pt}
    \begin{subfigure}[t]{0.20\textwidth}
        \centering
        \includegraphics[width=\linewidth,height=0.85\linewidth]{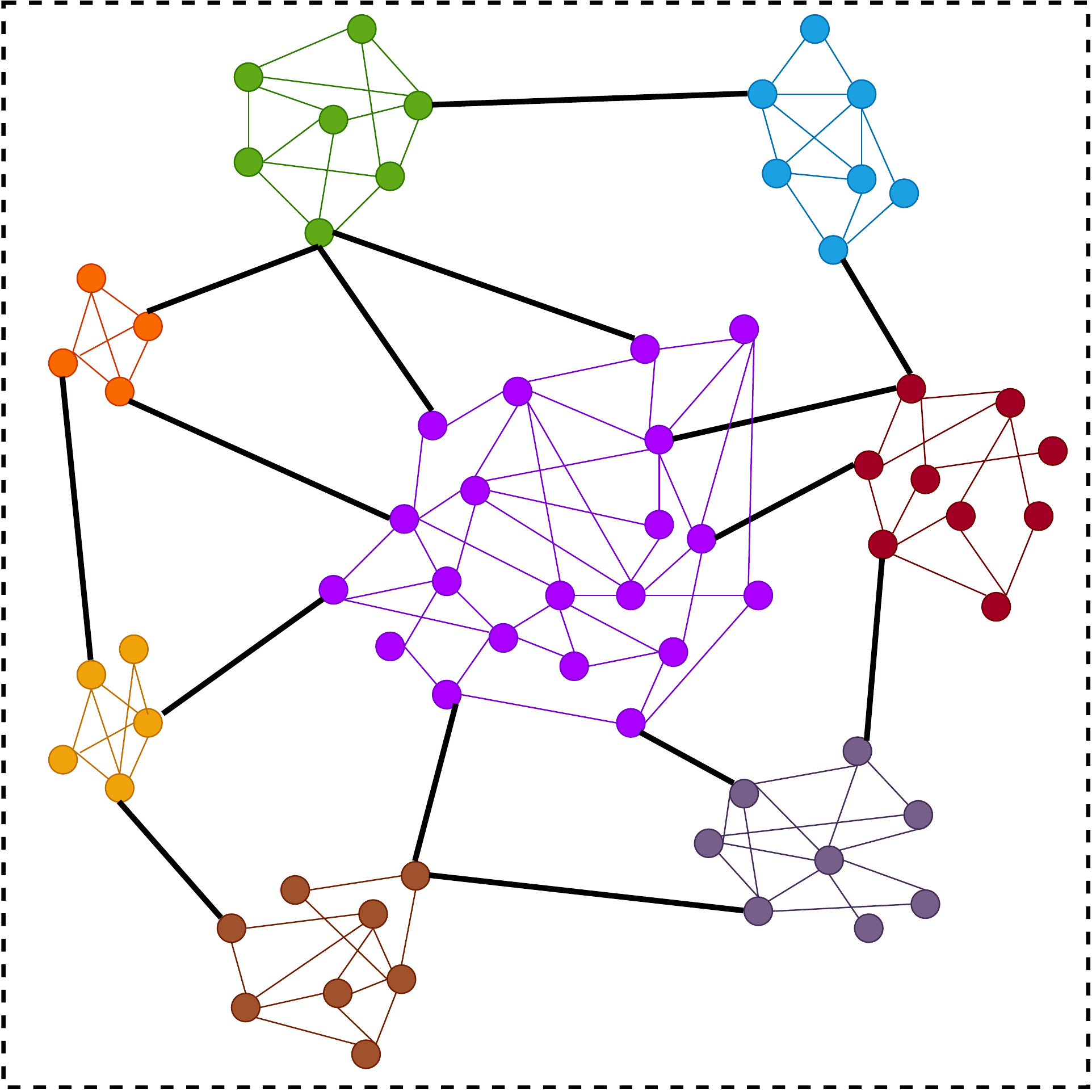}
        % \caption{Communities in Social Network}
        \caption{Network of Networks}
        \label{fig:social_network}
    \end{subfigure}
    % \hspace{23pt}
    % \begin{subfigure}[t]{0.25\textwidth}
    %     \centering
    %     \includegraphics[width=\linewidth]{figures/network_of_networks.pdf}
    %     \caption{Network of Networks}
    %     \label{fig:network_of_networks}
    % \end{subfigure}
    \caption{Examples of Interdependent Networks. (a) Temporal Networks: black dashed lines capture implicit temporal dependencies or consistencies. (b) Network of Networks: black solid lines are bridges across networks.}
    % (c) Network of Networks with Implicit Connections: the main network is represented by dashed nodes and edges, where each node of the main network can be represented as another network. Black dashed lines model the implicit connections.}
    \label{fig:interdependent_networks}
\end{figure}

Subgraph detection in network data has aroused many interests in recent years because of many real-world applications, such as disease outbreak detection \cite{chen2014non}, intrusion detection in computer networks %\cite{shao2018efficient}
, event detection in social networks \cite{rozenshtein2014event}, congestion detection in traffic networks, etc. However, most of existing works investigate the subgraph mining on static, isolated networks, and such a problem involving interdependent networks has not been well studied. 
% In interdependent networks, one network depends on another to achieve its full functionality. Examples include smart power grids, transportation networks, and layered communication networks \cite{danziger2014introduction}. 
Interdependent networks are comprised of multiple networks $\{\mathbb{G}^1, \mathbb{G}^2, \dots, \mathbb{G}^k,\dots\}$ and edges $\mathbb{E}^0$ interconnected among networks, where $\mathbb{G}^k = (\mathbb{V}^k, \mathbb{E}^k)$. $\mathbb{V}^k$ and $\mathbb{E}^k$ are vertex set and edge set of $k^{\text{th}}$ network $\mathbb{G}^k$ respectively. Some nodes in different networks exhibit node-node dependencies that could be captured by explicit edges or implicit correlation on node attributes (implicit edges). 
For instance, a temporal network can be viewed as multiple temporal-dependent networks, in which each network represents a snapshot of the temporal network at a specific time stamp, where current node's attributes depend on attributes in the previous time-stamp implicitly \cite{mucha2010community} (Figure \ref{fig:temporal_network}).  A web-scale social network comprised of many communities is a network of networks (a trivial interdependent networks) with explicit connections, where communities can be viewed as small networks or blocks that interconnect with each other (Figure \ref{fig:social_network}). 
% Another example of network of networks with implicit connections is infrastructure networks (Figure \ref{fig:network_of_networks}). 

% \textcolor{red}{ In reality, diverse critical infrastructures are coupled together and depend on each other, including systems such as water and food supply, communications, fuel, financial transactions and power stations. For instance, the electric power network provides power for pumping and control systems of water network, the fuel network provides fuel for generators for electric power network, etc \cite{gao2014single}.} 

% \textcolor{red}{Typically, subgraph detection in a single network is to find a subset of vertices and edges that minimizing an user-specified cost function, meanwhile, satisfying a certain topological structured constraint, such as connected subgraphs, dense subgraphs, subgraphs that are isomorphic to a query graph, compact subgraphs, trees, and paths, among others \cite{chen2016generalized}. It can be formulated as an NP-hard combinatorial optimization problem as follows:
% \begin{equation}\label{prob:single}
%      \begin{gathered}
%         \min_{S\subseteq \mathbb{V}} F(S) \\
%         \mathrm{s.t.} \ \ S \ \text{satisfies a pre-defined topological constraint}
%     \end{gathered}
% \end{equation}
% where $F$ is a user-specified cost function, and $S$ is a subset of vertex set $\mathbb{V}$ of a network. Correspondingly,} 

Subgraph detection in multiple interdependent networks can be formulated as a block-structured optimization problem with multiple topological constraints on blocks,
\begin{equation}\label{prob:discrete}
     \begin{gathered}
        \min_{S_1 \subseteq \mathbb{V}^1, ..., S_K \subseteq \mathbb{V}^K } F(S_1,\cdots, S_K) \\
        \mathrm{s.t.} \ \ S_k \ \text{satisfies a pre-defined topological constraint,}
    \end{gathered}
\end{equation}
where $F$ is a user-specified cost function regularized by block dependencies, for example, $F$ could be $f(S_1, \cdots, S_K) + g(S_1, \cdots, S_K)$, where $f$ is used to capture signals in interdependent networks and $g$ models the dependencies between networks. $S_k$ is a subset of nodes in $k^{\text{th}}$ network $\mathbb{G}^k$, $ k=1,...,K $. 
% \textcolor{red}{For example, subgraph detection in a temporal network is to find a sequence of subset of vertices in a sequence of blocks, where the detected subgraph in each block must satisfy a predefined topological constraint. }
Vanilla subgraph detection problem  
%(\ref{prob:single}) 
is a special case of problem (\ref{prob:discrete}) when number of networks (blocks) is 1. 

% There are many trivial examples of  multiple interdependent networks \cite{kivela2014multilayer,gao2012networks}. For instance, a temporal network can be viewed as multiple temporal-dependent networks, in which each network represents a snapshot of the temporal network at a specific time stamp \cite{mucha2010community}. Subgraph detection in a temporal network is to find a sequence of subset of vertices in a sequence of blocks, where the detected subgraph in each block must satisfy a predefined topological constraint. A large network comprised of many communities is a network of networks, where communities can be viewed as small networks or blocks that interconnect with each other. Vanilla subgraph detection problem is a special case of discovering subgraphs in interdependent networks when number of interdependent networks (blocks) is 1.

%  that maximize a score function $f({\bf x})$ with ${\bf x} \in \mathbb{R}^N$ associated with $N$ nodes, subject to a certain topological constraint. 
 
%  However, in many scenarios, we would like multiple networks satisfy different topological constraints, or 
% As one of major tasks in network mining, subgraph discovery in a single network has been well-studied, and the subgraph discovery among multiple interdependent networks becomes more and more popular research topic in recent years.

% \textcolor{red}{Subgraph detection in interdependent networks is very challenging but useful, since real systems are more often than not interconnected, with many inter-dependencies that are not properly captured by an isolated network. It is important to take such multiple networks which are interdependent with each other into account to try to improve our understanding of interesting patterns in networks.} 

To the best of our knowledge, most of related studies on subgraph detection in interdependent networks only focus on specific applications and are lack of generality. Furthermore, they are heuristic-driven with no theoretical guarantee. Therefore, we propose a general framework that leverages graph structured sparsity model \cite{hegde2015nearly} and block coordinate descent method \cite{fercoq2015accelerated} to solve this problem which can be modeled as a block-structured optimization problem.

The contributions of our work are summarized as follows:
\begin{itemize} [wide=0pt,leftmargin=\parindent]
    \item \textbf{Design of an efficient and scalable approximation algorithm.} We propose a novel generic framework, namely, Graph Block-structured Gradient Projection, for block structured nonconvex optimization, which can be used to approximately solve a broad class of  subgraph detection problems in interdependent networks in nearly-linear time. 
    % \textcolor{red}{Furthermore, we design a parallel version of our algorithm to make it more scalable and efficient. }
    \item \textbf{Theoretical guarantees.} We present a theoretical analysis of the proposed GBGP algorithm and show that it enjoys a good convergence rate and a tight error bound on the quality of the detected subgraph. 
    \item \textbf{Two practical applications with comprehensive experiments.} We demonstrate how our framework can be applied to two practical applications: 1) anomalous evolving subgraph detection; 2) subgraph detection in network of networks. We conduct comprehensive experiments on both synthetic and real networks to validate the effectiveness and efficiency of our proposed algorithm. 
\end{itemize}
% \textcolor{red}{\textbf{Reproducibility}: The implementation of GBGP, datasets, and supplementary material are available via the anonymous link: \url{https://bit.ly/2NxDvkw}}

% \textcolor{red}{The rest of this paper is organized as follows. Section \ref{section:method} introduces some preliminaries and the proposed method GBGP. Section \ref{section:theory} analyzes theoretical properties of the method. Section \ref{section:example} demonstrates applications of our proposed algorithm. Experimental design, datasets and results are shown in section \ref{section:experiment}. We introduce related work in Section \ref{section:related}. Section \ref{section:conclusion} concludes the paper and describes future work.}

\section{Methodology}\label{section:method}
\subsection{Problem Formulation}
% Firstly, multiple interdependent networks can be viewed as one combined network. 
First, we reformulate the combinatorial problem (\ref{prob:discrete}) in discrete space as an nonconvex optimization problem in continuous space. 
% \textcolor{red}{Interdependent networks can be viewed as one large network $\mathbb{G}= \{\mathbb{V}^1, \cdots, \mathbb{V}^K, \mathbb{E}^0, \mathbb{E}^1, \cdots, \mathbb{E}^K \}$, where each pair of $(\mathbb{V}^k, \mathbb{E}^k)$ forms a small network $\mathbb{G}^k$ for $k= 1, \cdots, K$, and $\mathbb{E}^0$ are interconnected edges across different small networks. Edges in $\mathbb{E}^0$ should be treated differently with the edges in each $\mathbb{E}^k$, since they models the dependencies among different networks.}
Interdependent networks can be viewed as one large network $\mathbb{G}= (\mathbb{V}, \mathbb{E})$, where $\mathbb{V} = \{1,\cdots, N\}$ could be cut into $\{\mathbb{V}^1, \cdots, \mathbb{V}^K \}$ and $\mathbb{E}$ could be split into $\{\mathbb{E}^0, \mathbb{E}^1, \cdots, \mathbb{E}^K\}$. Each pair of $(\mathbb{V}^k, \mathbb{E}^k)$ forms a small network $\mathbb{G}^k$ for $k = 1,\cdots, K$, and $\mathbb{E}^0$ are edges interconnected among different small networks. Edges in $\mathbb{E}^0$ should be treated differently with the edges in each $\mathbb{E}^k$, since they models the dependencies among different networks.
% Given interdependent networks $ \{\mathbb{G}^{1},\dots,\mathbb{G}^{K}\}$ where $ \mathbb{G}^{k} = (\mathbb{V}^{k}, \mathbb{E}^{k})$, $k=1,\dots,K$, the vertex set of all networks is denoted as $\mathbb{V} = \{1, \cdots, N\}$. It can be easily obtained that $\mathbb{V} = \mathbb{V}^1\cup  \cdots \cup \mathbb{V}^K$.  
${\bf W} = [{\bf  w}_1, \cdots, {\bf w}_N]\in \mathbb{R}^{P\times N}$ is the feature matrix, and ${\bf w}_i\in \mathbb{R}^P$ is the feature vector of vertex $i$, $i\in\mathbb{V}$. $N_k = |\mathbb{V}^k|$ is the size of the subset of vertices $\mathbb{V}^k$.   

The general subgraph detection problem in interdependent networks can be formulated as following general block-structured optimization problem with topological constraints: 
\begin{equation}
\begin{gathered}
\min_{{\bf x}=({\bf x}^1,\dots,{\bf x}^K)} F({\bf x}_1,\dots,{\bf x}_K) \ \  \\
\mathrm{s.t.}\ \ \mathrm{supp}({\bf x}^k) \in \mathbb{M}(\mathbb{G}^k, s),\ \  k = 1, \cdots, K \label{eq:gen_prob}
\end{gathered}
\end{equation}
where the vector ${\bf x} \in \mathbb{R}^N$ is partitioned into multiple disjoint blocks ${\bf x}^1 \in \mathbb{R}^{N_1}, \cdots, {\bf x}^K \in \mathbb{R}^{N_K}$, and ${\bf x}^k$ are variables associated with nodes of network $\mathbb{G}^k$. The objective function $F(\cdot)$ is a continuous, differentiable and convex function, which will be defined based on the feature matrix ${\bf W}$. In addition, $F(\cdot)$ could be decomposed as $f({\bf x}) + g({\bf x})$, where $f$ is used to capture signals on nodes in interdependent networks and $g$ models the dependencies between networks. $\text{supp}({\bf x}^k)$ denotes the support set of vector ${\bf x}^k$, $\mathbb{M}(\mathbb{G}^k, s)$
denotes all possible subsets of vertices in $\mathbb{G}^k$ that satisfy a certain predefined topological constraint. One example of topological constraint for defining $\mathbb{M}(\mathbb{G}^k, s)$ is connected subgraph, and we can formally define it as follows:
\begin{equation}
    \mathbb{M}(\mathbb{G}^k, s) \coloneqq \{S | S \subseteq \mathbb{V}^k; |S| \leq s; \mathbb{G}^k_S\ \text{is connected.} \}
\end{equation}
where $s$ is a predefined upperbound size of $S$, $S\subseteq \mathbb{V}^k$ , and $\mathbb{G}^k_S$ refers to the induced subgraph by a set of vertices $S$. The topological constraints can be any graph structured sparsity constraints on $\mathbb{G}^k_S$, such as connected subgraphs, dense subgraphs, compact subgraphs \cite{chen2016generalized}. Moreover, we do not restrict all $\mathrm{supp}({\bf x}^1),\cdots, \mathrm{supp}({\bf x}^K)$ satisfying an identical topological constraint. 
%\textcolor{red}{An illustration of problem formulation for connected subgraph detection in interdependent networks can be found in Figure \ref{fig:prob_demo}.}
% \setlength{\textfloatsep}{5pt}
% \begin{figure}[htbp]
%     \centering
%     \includegraphics[width=0.95\linewidth]{figures/prob_demo_4.pdf}
%     \caption{Illustration of connected subgraph detection in interdependent networks. In this example, interdependent networks $\{\mathbb{G}^1,\mathbb{G}^2,\mathbb{G}^3\}$ with an obvious block-separated structure are given. Dashed lines represent the connections across different networks, and solid lines represent connections between nodes in the same network. Those two types of lines may characterize different relationships in practical applications. Those red nodes are what we are interested in, and their corresponding entries in vector ${\bf x}$ should be nonzero, while others should be 0s once we get an optimum of ${\bf x}$.}
%     \label{fig:prob_demo}
% \end{figure}

% \noindent Notice that the objective function of Problem (\ref{eq:gen_prob}) is convex, while it has nonconvex constraints, which makes it a \textbf{nonconvex optimization problem}.

\begin{algorithm}%[H]
\begin{algorithmic}[1]
\Statex \textbf{Input}: $\{\mathbb{G}^{1},\dots,\mathbb{G}^{K}\}$
\Statex \textbf{Output}: ${\bf x}^{1, t}, \cdots, {\bf x}^{K, t}$
\Statex \textbf{Initialization}, $ i=0 $, $ {\bf x}^{k,i}=\text{initial vectors}$, k=1,\dots, K
\Repeat
    \For{$k = 1,\cdots, K$}
        \State $\Gamma_{{\bf x}^{k}}=H(\nabla_{{\bf x}^{k}} F({\bf x}^{1,i}, \dots,{\bf x}^{K,i}))$
        \State $\Omega_{{\bf x}^{k}}=\Gamma_{{\bf x}^{k}}\cup\mathrm{supp}({\bf x}^{k,i})$
    \EndFor
    % \abovedisplayskip=-0.7\baselineskip
    \State Get $({\bf b}_{{\bf x}^1}^i,\dots,{\bf b}_{{\bf x}^K}^i)$ by solving problem (\ref{sub-p1})\label{alg:line:subproblem}
    %\begin{flalign*}
    %     &\left( {\bf b}_{{\bf x}^1}^i,\dots,{\bf b}_{{\bf x}^K}^i \right)&\\
    %     &= \argmax\limits_{{\bf x}^1,\dots,{\bf x}^K} f({\bf x}^1,\dots,{\bf x}^K) + \sum_{k=1}^{K} g_k({\bf x}^k) & \\
    %     &\, \mathrm{s.t.}\,\, \mathrm{supp}({\bf x}^k)\subseteq \Omega_{{\bf x}^k}&
    % \end{flalign*}
    % \vspace{-\baselineskip}
    \For{$k = 1, \cdots, K$}
        \State $ \Psi_{{\bf x}^{k}}^{i+1}=T({\bf b}_{{\bf x}^k}^i) $
        \State $ {\bf x}^{k,i+1}=[{\bf b}_{{\bf x}^{k}}^i]_{\Psi_{{\bf x}^{k}}^{i+1}} $
    \EndFor
    \State $ i=i+1 $
\Until{$ \sum_{k=1}^{K} \left\|{\bf x}^{k,i+1}-{\bf x}^{k,i}\right\|\leq \epsilon $}
\State $ C=(\Psi_{x^1}^{i}, \dots, \Psi_{{\bf x}^k}^{i}) $
\Return $ ( {\bf x}^{1,i}, \cdots, {\bf x}^{K,i}) , C $
\end{algorithmic}
\caption{Graph Block-structured Gradient Projection}\label{alg:gbmp}
\end{algorithm}
\setlength{\textfloatsep}{7pt} % for single column

\subsection{Head and Tail Projections on $\mathbb{M}(\mathbb{G}, s)$}
% \textcolor{red}{Before presenting our algorithm, we first introduce two major components related to the support of the topological constraints "$\text{supp}(\bf{x}) \in \mathbb{M}(\mathbb{G}, s)$", including head and tail projections \cite{hegde2015nearly}. The key idea is that, suppose we are able to find a good intermediate solution $\bf{x}$ that does not satisfy this topological constraint, these two types of projections can be used to find good \textit{approximations} of $\bf{x}$ in the feasible space defined by $\mathbb{M}(\mathbb{G},s)$.}

\begin{itemize}[leftmargin=*]
    \item \textbf{Tail Projection} ($T({\bf x})$): is to find a subset of nodes $S \subseteq \mathbb{V}$ such that 
    \begin{equation}
        \|{\bf x} - {\bf x}_{S}\|_2 \leq c_{T} \cdot \min_{S' \in \mathbb{M}(\mathbb{G}, s)} \|{\bf x} - {\bf x}_{S'}\|_2,
    \end{equation}
    where $c_{T} \geq 1$, and ${\bf x}_S$ is a restriction of ${\bf x}$ on $S$ such that: $({\bf x}_S)_i = ({\bf x})_i$ if $i \in S$, and $({\bf x}_S)_i=0$ otherwise. When $c_{T} = 1$, $T({\bf x})$ returns an optimal solution to the problem: $\min_{S'\in \mathbb{M}(\mathbb{G},s)} \|{\bf x}-{\bf x}_{S'}\|_2$. When $c_T > 1$, $T({\bf x})$ returns an approximate solution to this problem with the approximation factor $c_T$.
    
    \item \textbf{Head Projection} ($H({\bf x})$): is to find a subset of nodes $S$ such that 
    \begin{equation}
        \|{\bf x}_{S}\|_2 \geq c_{H}\cdot \max_{S'\in \mathbb{M}(\mathbb{G}, s)} \|{\bf x}_{S'}\|_2,
    \end{equation}
    where $c_H\leq1$. When $c_H = 1$, $H({\bf x})$ returns an optimal solution to the problem: $\max_{S' \in \mathbb{M}(\mathbb{G},s)} \|{\bf x}_{S'}\|_2$. When $c_H < 1$, $H({\bf x})$ returns an approximate solution to this problem with the approximation factor $c_H$.
\end{itemize}

% \textcolor{red}{It can be readily proved that, when $c_T = 1$ and $c_H = 1$, both $T({\bf x})$ and $H({\bf x})$ return the same subset $S$, since $\min_{S'\in \mathbb{M}(\mathbb{G},s)} \|{\bf x}-{\bf x}_{S'}\|_2  = \max_{S' \in \mathbb{M}(\mathbb{G},s)} \|{\bf x}_{S'}\|_2$. In addition, the corresponding vector ${\bf x}_S$ is an optimal solution to the standard projection oracle in the traditional projected gradient descent algorithm \cite{bahmani2016learning}:
% \begin{equation}
%     \argmin_{{\bf x}'\in \mathbb{R}^n} \|\bf{x}-\bf{x}'\|_2 \ \ \text{s.t.\ supp}({\bf x}') \in \mathbb{M}(\mathbb{G}, s) \label{eq:projection_oracle}
% \end{equation}
% which is NP-hard in general for popular topological constraints, such as connected subgraphs and dense subgraphs \cite{qian2014connected}. However, when $c_T > 1$ and $c_H < 1$, $T({\bf x})$ and $H({\bf x})$ return different approximate solutions to the standard projection problem \cite{chen2017generic}.}

Although the head and tail projections are NP-hard when we restrict $c_T = 1$ and $c_H = 1$, these two projections can still be implemented in nearly-linear time when approximated solutions with $c_{T} > 1$ and $c_H < 1$ are allowed. 
% \textcolor{red}{The head and tail projections are two different approximations to the standard projection problem (\ref{eq:projection_oracle}). It has been demonstrated that the joint utilization of both head and tail projections is critical in design of approximate algorithms for network topology-related optimization problems \cite{chen2016generalized,hegde2015nearly,hegde2015approximation,zhou2016graph}.}

\subsection{Algorithm Details}
We propose a novel Graph Block-structured Gradient Projection, namely GBGP, to approximately solve problem (\ref{eq:gen_prob}) in nearly-linear time on the network size. The key idea is to alternatively search for a close-to-optimal solution by solving easier sub-problems for graph $\mathbb{G}_k$ in each iteration $i$ until converged. The pseudo-code of our proposed algorithm is described in Algorithm \ref{alg:gbmp}. Our algorithm can be decomposed into three main steps, including:
\begin{itemize}[leftmargin=*]
    \item \textbf{Step 1}: alternatively identify a subset of nodes in each block $\Omega_{{\bf x}^k}$, in which pursuing the minimization will be most effective (\textbf{Line 2 $\sim$ 5}).
    \item \textbf{Step 2}: identify the intermediate solution $({\bf b}_{{\bf x}^1}^i,\dots,{\bf b}_{{\bf x}^K}^i)$ that minimizes the objective function in intermediate space $\cup_{k=1}^{K} \Omega_{{\bf x}^k}$ (\textbf{Line 6});
    \begin{equation}
    \begin{gathered}
      \hspace*{-10pt}\left( {\bf b}_{{\bf x}^1}^i,\dots,{\bf b}_{{\bf x}^K}^i \right) = \argmin\limits_{{\bf x}^1,\dots,{\bf x}^K} F({\bf x}^1,\dots,{\bf x}^K) \\
      \mathrm{s.t.}\, \ \ \mathrm{supp}({\bf x}^k)\subseteq \Omega_{{\bf x}^k} \label{sub-p1}
    \end{gathered}
    \end{equation}
    \item \textbf{Step 3}: alternatively apply tail projections on the intermediate solution $({\bf b}_{{\bf x}^1}^i,\dots,{\bf b}_{{\bf x}^K}^i)$ to the feasible space defined by constraints: ``$\text{supp}({\bf x}^k) \in \mathbb{M}(\mathbb{G}^k, s)$" (\textbf{Line 7 $\sim$ 10}).
\end{itemize}

We utilize the \textit{block-coordinate descent method} with \textit{proximal linear update} \cite{tseng2009coordinate,shi2016primer} to solve the problem (\ref{sub-p1}) (Algorithm \ref{alg:block}). 
% \textcolor{red}{This method has been analyzed and also applied to both convex and nonconvex problems \cite{beck2013convergence,zhou2016global,hong2017iteration,xu2017globally}, which shows good performance empirically. Block coordinate descent is a generalization of the alternating minimization method that has been applied to a variety of problems, such as the expectation-maximization (EM) algorithm \cite{dempster1977maximum}.} 
In addition, proximal linear update is used to ensure the convergence of the algorithm on convex problems with convex constraints ``$\textrm{supp}({\bf x}^{k}) \subseteq \Omega_{{\bf x}^k}$". The proximal linear update in our scenario is defined by:
\begin{equation}\label{prob:proximal}
{\fontsize{9}{10}
\begin{aligned}
\hspace{-10pt}{\bf x}^{k, t+1} = &\argmin_{{\bf x}^k}  F(\hat{\bf x}^{t}) + \langle \nabla_{{\bf x}^k} F(\hat{\bf x}^{k,t}, \hat{\bf x}^{\neq k, t}), {\bf x}^{k} - \hat{\bf x}^{k, t} \rangle\\
& + \frac{1}{2 \alpha^{k, t}}\|{\bf x}^{k} - {\bf \hat{x}}^{k, t}\|_2^2\qquad\mathrm{s.t.} \ \ \mathrm{supp}({\bf x}^{k}) \subseteq \Omega_{{\bf x}^k}\hspace{-7pt}
\end{aligned}
}
\end{equation}
% \begin{equation}\label{prob:proximal}
% \begin{split}
% \hspace{-8pt}{\bf x}^{k, t+1} = &\argmin_{{\bf x}^k}  F(\hat{\bf x}^{t}) + \langle \nabla_{k} F(\hat{\bf x}^{k,t}, \hat{\bf x}^{\neq k, t}), {\bf x}^{k} - \hat{\bf x}^{k, t} \rangle\\
% & + \frac{1}{2 \alpha^{k, t}}\|{\bf x}^{k} - {\bf \hat{x}}^{k, t}\|_2^2 + h_k({\bf x}^k)
% \end{split}
% \end{equation} 
% where $h_k({\bf x}^k)$ is the indicator function of the convex set constraint ``$\textrm{supp}({\bf x}^{k}) \subseteq \Omega_{{\bf x}^k}$", 
% We subtlety abuse the notation in algorithm (\ref{alg:block}) for more clear explanation, in where ${\bf x}_i^t$ denotes by the $i$-th block of ${\bf x}$ at $t$-th iteration. 
where $\alpha^{k, t}$ serves as a step size and can be set as the reciprocal of the Lipschitz constant of $\nabla_{k} F(\hat{\bf x}^{k,t}, \hat{\bf x}^{\neq k, t})$, and $\hat{\bf x}^{k, t}$ (\textbf{Line 4}) is an extrapolated point that helps accelerate the convergence of the proximal point update scheme. 
% \begin{equation}
%   \mathrm{Prox}_{\eta, h}( {\bf \hat{x}}^{k, t} - \frac{1}{\alpha^{k, t}} \nabla_{k} F(\hat{\bf x}^{k,t}, \hat{\bf x}^{\neq k, t}) ) =  
% \end{equation}
% \textcolor{red}{Since the objective function is simply second order Taylor approximation of function $F$ with Hessian replaced by identity matrix, we can easily derive and implement the closed form solution of the objective function in problem (\ref{prob:proximal}), and then project it to the feasible space, which is convex.} 
The overall block coordinated gradient projection method on convex function with convex constraint (i.e. Algorithm \ref{alg:block}) has a sublinear rate of convergence \cite{shi2016primer}. 
\begin{algorithm}[H]
\begin{algorithmic}[1]
\Statex \textbf{Input}: $\{\mathbb{G}^{1},\dots,\mathbb{G}^{K}\}$
\Statex \textbf{Output}: ${\bf x}^{1, t}, \cdots, {\bf x}^{K, t}$
\Statex \textbf{Initialization}: $t = 0, \epsilon = 10^{-3}, \rho_0 = 1.$
\Repeat
    \State \text{Choose index } $k \in \{1, \cdots, K\}$
    \State $\omega_{t} = (\rho_t - 1) / \rho_t,\ $
    \State $\hat{\bf x}^{k, t} = {\bf x}^{k, t} + \omega_{t} ({\bf x}^{k, t} - {\bf x}^{k,t-1})$
    % \State Update ${\bf x}^{k, t}$ to ${\bf x}^{k, t+1}$ by solving Problem (\ref{prob:proximal})
    \State Update ${\bf x}^{k, t + 1} \leftarrow {\bf \hat{x}}^{k, t} - \frac{1}{\alpha^{k, t}} \nabla_{k} F(\hat{\bf x}^{k,t}, \hat{\bf x}^{\neq k, t})$ 
    \State Project ${\bf x}^{k, t+1}$ to feasible space by setting entries of ${\bf x}^{k, t+1}$ to zero if index of entry not in set $\Omega_{{\bf x}^k}$. 
    \State Keep ${\bf x}^{j, t+1} = {\bf x}^{j, t}$, for all $j \neq k$
    \State $\rho_{t+1} = (1+\sqrt{1 + 4 \rho_{t}^2})/2,$
    \State Let $t = t+1$
\Until{$\sum_{k=1}^{K} \|{\bf x}^{k, t} - {\bf x}^{k, t-1}\| \le \epsilon$}
\Return $\{{\bf x}^{1, t}, \cdots, {\bf x}^{K, t}\}$
\end{algorithmic}
\caption{Block-Coordinate Descent Method with Proximal Linear Update to Solve Problem (\ref{sub-p1})}\label{alg:block}
\end{algorithm}
\section{Theoretical Analysis} \label{section:theory}
In order to demonstrate the accuracy and efficiency of GBGP, we require that the objective function $F({\bf x})$ satisfies the Weak Restricted Strong Convexity (WRSC) condition, which is a variant of the Restricted Strong Convexity/Smoothness (RSC/RSS) \cite{yuan2014gradient}:

\begin{definition}[Weak Restricted Strong Convexity (WRSC)]\label{def:wrsc} A function $F({\bf x})$ has condition $(\xi,\delta,\mathbb{M})$-WRSC, if $\forall {\bf x}, {\bf y}\in\mathbb{R}^N$ and $\forall S\in\mathbb{M}$ with $\mathrm{supp}({\bf x})\cup\mathrm{supp}({\bf y})\subseteq S$, the following inequality holds for some $\xi >0$ and $0<\delta<1$:
\begin{equation}
    \|{\bf x}-{\bf y}-\xi\nabla_S F({\bf x})+\xi\nabla_S F({\bf y})\|_2\leq\delta\|{\bf x}-{\bf y}\|_2
\end{equation}
where ${\bf x}=({\bf x}^1,\dots,{\bf x}^K), {\bf y}=({\bf y}^1,\dots,{\bf y}^K), {\bf x}^k, {\bf y}^k\in\mathbb{R}^{N_k}, k=1,\dots,K$,  topological constraint $\mathbb{M}$ can be expressed as $\mathbb{M}(\mathbb{G}, s)=\bigcup_{k=1}^K\mathbb{M}(\mathbb{G}^k,s_k), s=\sum_{k=1}^K s_k$, and the subgraph in $k^{\text{th}}$ block (i.e., $\mathbb{G}^k$)  is $S_k$, which satisfies $|S_k|\leq s_k, S_k\subseteq \mathbb{V}^k, S=\bigcup_{k=1}^K S_k, |S|\leq s$. Here, since constraints on blocks are independent, we use union sign ``$\bigcup$" to denote combined model $\mathbb{M}$, in which ${\bf x}\in\mathbb{M}=\{{\bf x}|{\bf x}^k\in\mathbb{M}(\mathbb{G}^k,s_k),k=1,\dots,K\}$.
\end{definition}
% \textcolor{red}{
% \begin{remark}
% We can set different $s_k$ for $k^{\text{th}}$ block. In our applications, $s_1=\cdots=s_k=s'$, i.e., we use the same upper bound of subgraph size for all blocks.
% \end{remark}}

\begin{theorem}\label{thm:convege}
Consider the graph block-structured constraint with $K$ blocks $\mathbb{M}(\mathbb{G}, s)=\bigcup_{k=1}^K\mathbb{M}(\mathbb{G}^k,s_k)$ and a cost function $F:\mathbb{R}^N\to\mathbb{R}$ that satisfies condition $(\xi,\delta,\mathbb{M}(\mathbb{G},8s))$-WRSC. If $\eta=c_H(1-\delta)-\delta>0$, then for any true ${\bf x}^*\in\mathbb{R}^N$ with $\mathrm{supp}({\bf x}^*)\in\mathbb{M}((\mathbb{G},s)$, the iteration of algorithm obeys
\begin{equation}
    \|{\bf x}^{i+1}-{\bf x}^*\|_2\leq\alpha\|{\bf x}^i-{\bf x}^*\|_2+\beta\|\nabla_I F({\bf x}^*)\|_2
\end{equation}
where $c_H=\min_{k=1,\dots,K}\{{c_{H_k}}\}$, $c_T=\max_{k=1,\dots,K}\{{c_{T_k}}\}$, $I=\argmax_{S\in\mathbb{M}}\|\nabla_S F({\bf x})\|_2$, $\alpha=\frac{1+c_T}{1-\delta}\sqrt{1-\eta^2}$, and $\beta=\frac{\xi(1+c_T)}{1-\delta}\left[\frac{1+c_H}{\eta}+\frac{\eta(1+c_H)}{\sqrt{1-\eta^2}}+1 \right]$. $c_{H_k}$ and $c_{T_k}$ denote head and tail projection approximation factors on $k^{\text{th}}$ block.
\end{theorem}
% \textcolor{red}{
% \begin{IEEEproof}
% The proof is provided in Appendix \ref{proof:converge}.
% \end{IEEEproof}
% \begin{remark}
% 1) The convergence of GBGP is controled by the shrinkage rate $\alpha<1$, which can be satisfied if and only if $c_H^2>1-1/(1+c_T)^2$ when $\delta$ is small. As proved in \cite{hegde2015approximation}, the approximation factor $c_H$ of any given head approximation algorithm can be boosted to any arbitrary constant close to 1, such that the above condition is satisfied. 2) According to WRSC condition, the function is Lipschitz continuous, which further implies the $\|\nabla_I F(x^*)\|_2$ must be bounded. To sum up, Theorem \ref{thm:convege} indicates the convegence of GBGP algorihtm.
% \end{remark}}

\begin{theorem}\label{thm:time}
Let ${\bf x}^*\in\mathbb{R}^N$ be a true optimum such that $\mathrm{supp}({\bf x}^*)\in\mathbb{M}(\mathbb{G}, s)$, and $F:\mathbb{R}^N\to\mathbb{R}$ be a cost function that satisfies condition $(\xi,\delta,\mathbb{M}(\mathbb{G}, 8s))$-WRSC. Assuming that $\alpha<1$, GBGP returns an $\hat{{\bf x}}$ such that, $\mathrm{supp}(\hat{{\bf x}})\in\mathbb{M}(\mathbb{G}, 5s)$ and $\|{\bf x}^*-\hat{{\bf x}}\|_2\leq c\|\nabla_{I}F({\bf x}^*)\|_2$, where $c=(1+\frac{\beta}{1-\alpha})$ is a fixed constant. Moreover, GBGP runs in time
\begin{equation}
{\fontsize{8}{8}
    O\left(\left(T+\sum_{k=1}^K|\mathbb{E}^k|\log^3 N_k\right)\log\left(\frac{\|{\bf x}^*\|_2}{\|\nabla_{I}F({\bf x}^*)\|_2}\right) \right)
}
\end{equation}
where $|\mathbb{E}^k|$, $N_k$ denote edge and node size of $k^{\text{th}}$ block and $T$ is the time complexity of one execution of the subproblem in line \ref{alg:line:subproblem} of Algorithm \ref{alg:gbmp}. In particularly, if $T$ scales linearly with $N$ and $|\mathbb{E}|$, then GBGP scales \textbf{nearly linearly} with $N$ and $|\mathbb{E}|$.
\end{theorem}

Note that the proofs of Theorem \ref{thm:convege} and Theorem \ref{thm:time} are omitted due to space limitation.

% \textcolor{red}{
% \begin{IEEEproof}
% The proof is provided in Appendix \ref{proof:time}.
% \end{IEEEproof}
% \begin{remark}
% We can run head and tail projections on blocks \textbf{in parallel}, which reduces the time cost of each iteration to $(T+|\mathbb{E}'|\log^3N'), |\mathbb{E}'|\log^3N'=\max_{k=1,\dots,K}|\mathbb{E}^k|\log^3N_k$. As mentioned in Section \ref{section:method}, which has been proved in \cite{beck2013convergence}, sublinear convergence rate $O(1/t)$ can be established for Algorithm \ref{alg:block}. Hence, the iteration number of Algorithm \ref{alg:block} is $O(\lceil 1/\epsilon\rceil)$ when error bound is $\epsilon$. Then it concludes that the time complexity of Algorithm \ref{alg:block} (i.e. $T$) is $O(\lceil 1/\epsilon\rceil N)$, which further 
% implies the algorithm GBGP scales nearly linearly with $N$ and $|\mathbb{E}|$.
% %According to \cite{fercoq2015accelerated}, solving subproblem (\ref{sub-p1}) enjoys complexity bound \textcolor{red}{ $O(1/t^2)$}, and complexity of each iteration is $O(N)$. Both of these two conclusions make the time complexity of Algorithm \ref{alg:block} and \ref{alg:approx_block} (i.e., $T$) become $O(\sqrt{\frac{1}{\epsilon}}N)$ when error bound is $\epsilon$, which scales linearly with $N$, and further implies GBGP algorithm scales nealy linearly with $N$.
% \end{remark}}

\section{Example Applications} \label{section:example}
% In this section, we show how to formulate two applications in the field of event detection: 1) anomalous evolving subgraph detection and 2) subgraph detection in network of networks, both of which could be well-formulated as problem (\ref{eq:gen_prob}).

In this section, we show how to formulate two subgraph detection applications: 1) anomalous evolving subgraph detection and 2) subgraph detection in network of networks as problem (\ref{eq:gen_prob}) with specific objective function $F$ and topological constraints. For these two applications, we leverage the \textbf{Elevated Mean Scan} (EMS) statistics, which is 
% popularly used for detecting signals among node-level numerical features on graph \cite{qian2014connected,arias2011detection}. The EMS statistics 
is defined as:
$ {\bf c}^\top {\bf x}/\sqrt{{\bf x}^\top {\bf 1}}$, where ${\bf x} \in \{0,1\}^N$, ${\bf c}$ denotes the feature vector of all nodes, and $c_i \in \mathbb{R}$ denotes the uni-variate feature for node $i$. Assuming $S$ is some unknown anomalous cluster which forms a connected component, $S\subseteq \mathbb{V}$. 
% The aim of significant subgraph detection is to decide between the null hypothesis $H_0 : \{ c_i \sim \mathbb{N}(0,1), \ \forall \ i \in V\}$ and the alternative hypothesis $H_{1, S}: \{c_i \sim \mathbb{N}(\mu, 1)\ \  \forall \ i \in S\ \text{with}\ \mu > 0 , \ \text{and}\ c_i \sim \mathbb{N}(0,1) \ \forall \ i \notin S\}$. Briefly, we hypothesize that the significant nodes $v \in S$ follow a different signal distribution with insignificant nodes $v \in V \setminus S$. 
Empirically, maximizing the score of EMS leads to discovering significant nodes in the network precisely. Instead of maximizing the EMS in the domain $\{0,1\}^N$, we relax EMS to continuous space and minimize the relaxed negative EMS in our applications, which can be defined as:
\begin{equation}
    -\frac{({\bf c}^\top {\bf x})^2}{ {\bf x}^\top {\bf 1}}  + \frac{1}{2} \|{\bf x}\|_2^2 \ \ \ \text{where} \ \ {\bf x} \in [0, 1]^N
\end{equation}
Most importantly, the relaxed negative EMS satisfies the RSC/RSS condition when ${\bf c}$ is normalized, which implies WRSC condition \cite{chen2016generalized,yuan2014gradient}.

\subsection{Anomalous Evolving Subgraphs Detection}
% Typically, an evolving event characterizes three phrases: emerging, spreading, and receding over a period of time. In order to detect and track the exact spatial and temporal region of an evolving event, we can formulate the problem as an anomalous evolving subgraph detection problem, which is to find a connected subgraph $S_k$ at each time stamp $k$ (denoted as \textit{local connectivity constraint}), and at the meanwhile we expect $S_k$ and $S_{k-1}$ share some overlap nodes in the network (denoted as \textit{temporal consistency constraint}). Hence, 
We can leverage the relaxed EMS and mathematically formulate the anomalous evolving subgraphs detection problem as nonconvex optimization with convex objective function and block-structured constraints:
\begin{equation}
\hspace{-5pt}
{\fontsize{8}{9}
\begin{aligned}
    \min_{{\bf x}^{1},\cdots,{\bf x}^{K}} &\sum_{k=1}^{K} \left(-\frac{ ({{\bf c}^k}^\top {\bf x}^k)^2}{{{\bf x}^k}^\top {\bf 1}} + \frac{1}{2}\|{\bf x}^{k}\|_2^2 \right)+ \lambda \cdot \sum_{k=2}^{K} \|{\bf x}^k - {\bf x}^{k-1}\|_2^2 \\
    &\ \ \ \ \ \ \ \ \ \ \ \mathrm{s.t.}\,\, \mathrm{supp}({\bf x}^k) \in \mathbb{M}(\mathbb{G}^k, s)
\end{aligned}\hspace{-10pt}
}
\end{equation}
where the first term is the summation of relaxed negative EMS, and the second term is soft constraints on ${\bf x}^{k}$ and ${\bf x}^{k-1}$ to ensure temporal consistency on detected subgraphs, and $\lambda > 0$ is a trade-off parameter. The connected subset of nodes at time stamp $k$ can be found as $S_k = \mathrm{supp}({\bf x}^{k})$, i.e., the support set of the estimated ${\bf x}^{k}$ that minimizes the objective function.

\subsection{Subgraph Detection in Network of Networks}
Our proposed framework is also applicable to subgraph detection in network of networks. 
% \textcolor{red}{Due to space limit, we only demonstrate the network of networks with explicit block dependencies, but our model can definitely apply to network of networks with implicit dependencies as well.} 
% A large-scale static network with many communities can be viewed as one instance of network of networks (a trivial interdependent networks), where each community is a small block of the network.  When an event is wide-spreading in such large networks, it becomes very challenging and time-consuming to apply effective models to detect the significant subgraphs. For example, one application is rumor detection and tracking in social networks. It is interesting and useful to identify the connected subgraph that depicts how a rumor spreads across different communities in a social network.
\begin{table*}[!htbp]
\caption{Results on synthetic datasets with different $\mu$. It shows that GBGP is more robust than Meden and Netspot.}\label{tab:noise}
\begin{adjustbox}{center}
\centering
\resizebox{0.75\textwidth}{!}{
\begin{tabular}{|c|c|c|c|c|c|c|c|c|c|}
\hline
\multirow{2}{*}{Methods} & \multicolumn{3}{c|}{$\mu=3$} & \multicolumn{3}{c|}{$\mu=4$} & \multicolumn{3}{c|}{$\mu=5$} \\ \cline{2-10}
 & \multicolumn{1}{c|}{Precision} & \multicolumn{1}{c|}{Recall} & F-measure & \multicolumn{1}{c|}{Precision} & \multicolumn{1}{c|}{Recall} & F-measure & \multicolumn{1}{c|}{Precision} & \multicolumn{1}{c|}{Recall} & F-measure \\ \hline
Meden & 0.7588 & 0.7342 & 0.7453 & 0.8836 & 0.8591 & 0.8709 & 0.9646 & 0.9145 & 0.9388 \\
NetSpot & 0.6658 & 0.7267 & 0.6947 & 0.7615 & 0.7922 & 0.7763 & 0.7956 & 0.8185 & 0.8068 \\
GBGP & 0.6468 & 0.8899 & \textbf{0.7489} & 0.8487 & 0.9674 & \textbf{0.9041} & 0.9553 & 0.9914 & \textbf{0.9730} \\ \hline
\end{tabular}
}
\end{adjustbox}
\end{table*}
% The most traditional approach is to partition a large network into several small blocks, and then process them individually and independently. However, this approach will effect the detection performance if those blocks are highly interdependent. By encoding the dependencies 
% in our proposed framework, we can detect subgraphs in each individual partition of networks more efficiently without sacrificing performance. 
% Specifically, our proposed framework provides a feasible solution for this scenario where node dependencies among a large-scale network can not be neglected. 
For subgraph detection in a network of networks, we can also leverage the relaxed negative $\text{EMS}$ and formulate the detection problem in large-scale networks as follows:
\vspace{-2pt}
\begin{equation}
\hspace{-3pt}
{\fontsize{8}{9}
    \begin{aligned}
        \min_{{\bf x}^{1},\cdots,{\bf x}^{K}} &\sum_{k=1}^{K} \left(-\frac{({{\bf c}^{k}}^\top {\bf x}^{k})^2}{{{\bf x}^{k}}^\top {\bf 1}} + \frac{1}{2}\|{\bf x}^{k}\|_2^2 \right)+\lambda \cdot \sum_{i,j} e_{ij} \cdot (x_i - x_j)^2 \\
        &\ \ \ \ \ \ \ \ \ \ \ \ \ \mathrm{s.t.}\,\, \mathrm{supp}({\bf x}^k) \in \mathbb{M}(\mathbb{G}^k, s)
	\end{aligned}
}\hspace{-15pt}
\end{equation}
where the first term is the summation of relaxed negative EMS, the second term is soft constraints on bridge nodes of two partitions to ensure dependencies; $e_{ij} = 1$ if node $i$ and node $j$ are connected but in two different partitions (in other words, edge $(i,j)$ is an graph cut), otherwise $e_{ij} = 0$, $x_i$ and $x_j$ are $i^{\text{th}}$ and $j^{\text{th}}$ entries of ${\bf x}$, and $\lambda > 0$ is a trade-off parameter. In addition, we propose a parallel version of our algorithm  to speed up the computation by integrating the APPROX algorithm, a randomized coordinate descent method proposed in \cite{fercoq2015accelerated}.

\section{Experiments}\label{section:experiment}
% This section evaluates the effectiveness and efficiency of the proposed GBGP algorithm in two applications on both synthetic data and real-world data.

\begin{table}[!htbp]
\caption{Statistics of Datasets for the 1st Application.}
\begin{adjustbox}{center}
\resizebox{0.85\linewidth}{!}{
\centering
\begin{tabular}{|c|c|c|c|c|}
\hline
\multirow{2}{*}{Datasets} & \multicolumn{4}{c|}{Statistics} \\ \cline{2-5} 
& \multicolumn{1}{c|}{Node} & \multicolumn{1}{c|}{Edge} & \multicolumn{1}{c|}{Timestamp} & \multicolumn{1}{c|}{Resolution} \\ \hline
Synthetic & 3,000 & 11,984 & 7 & NA \\
 Water Pollution & 12,527 & 14,831 & 8 & 60 min. \\
Washington D.C. & 1,188 & 1,323 & 17 & 60 min. \\
Beijing & 59,000 & 70,317 & 12 & 10 min. \\ \hline
\end{tabular}\label{table:app1}
}
\end{adjustbox}
\end{table}

\subsection{Anomalous Evolving Subgraph Detection}

\paragraph{Synthetic Dataset} We generate networks using Barab\'asi-Albert preferential attachment model \cite{barabasi1999emergence}. The evolving true subgraphs spanning within 7 time stamps are simulated from node size 100 to 300, and the true subgraphs in two consecutive time stamps have $50\%$ of node overlap. The univariate feature values of background nodes and true nodes are randomly generated in $\mathbb{N}(0,1)$ and $\mathbb{N}(\mu, 1)$ distributions, respectively. We generate 50 temporal networks for each setting of $\mu = [3,4,5]$. 
% As we can see, when $\mu$ is small, the data is more noised, and it is more difficult to distinguish between the anomalous nodes and normal nodes based on the univariate feature.

\paragraph{Real-world Dataset} 1) \textbf{\textit{ Water Pollution Dataset}}: a real world sensor network \cite{ostfeld2008battle}.
% with 12,527 vertices and 14,831 edges. 
% \textcolor{red}{This temporal network consists of node attributes in 8 time stamps with time resolution as 60 minutes, and the evolving true chemical contaminated subgraphs are provided.} 
For each hour, each vertex has a sensor that reports 1 if it is polluted; otherwise, reports 0.
2) \textbf{\textit{Washington D.C. Road Traffic Dataset}}: a traffic dataset of Washington D.C 
% from June 1, 2013 to March 31, 2014 
from INRIX \footnote{http://inrix.com/publicsector.asp.}.
%\cite{chen2017collective}
% The road networks consists of 1,188 nodes and 1,323 edges, and 
% We use the data from 6AM. to 10PM. (17 time stamps) with time resolution as 60 minutes. 
3) \textbf{\textit{ Beijing Road Traffic Dataset}}: the dataset contains the real-time traffic conditions of Beijing city. 
% The road networks consists of 59,000 nodes and 70,317 edges, and 
% We use the data from 5PM. to 7PM. with time resolution as 10 minutes (totally 12 time stamps) 
\cite{shang2014inferring}. For both traffic datasets, the node attribute is the difference between reference speed and current speed, and the true congested roads are provided. Statistics of all datasets are provided in Table \ref{table:app1}.

% \begin{table*}[ht]
% \caption{Results on Water Pollution, Washingtong D.C. and Beijing datasets. }\label{tab:evo}
% % \vspace{-3pt}
% \centering
% \setlength\tabcolsep{2.5pt}
% \resizebox{0.8\textwidth}{!}{
% \begin{tabular}{|c|c|c|c|c|c|c|c|c|c|}
% \hline
% \multirow{2}{*}{Methods} & \multicolumn{3}{c|}{Water Pollution} & \multicolumn{3}{c|}{Washington D.C.} & \multicolumn{3}{c|}{Beijing} \\ \cline{2-10}
%  & \multicolumn{1}{c|}{Precision} & \multicolumn{1}{c|}{Recall} & F-measure & \multicolumn{1}{c|}{Precision} & \multicolumn{1}{c|}{Recall} & F-measure & \multicolumn{1}{c|}{Precision} & \multicolumn{1}{c|}{Recall} & F-measure \\ \hline
% Meden &  0.8581 & 0.9330 & 0.8940 & 0.7076 & 0.7662 & 0.7342 & 0.6424 & 0.7509 & 0.6882 \\
% NetSpot & 0.9630 & 0.6591 & 0.7826 & 0.5823 & 0.7098 & 0.6367 & 0.6789 & 0.7351 & 0.6973 \\
% GBGP & 0.9206 & 0.8981 & \textbf{0.9092} &0.7049 & 0.9192 & \textbf{0.7853} & 0.6627 & 0.9634 & \textbf{0.7788} \\ \hline
% \end{tabular}
% }
% \end{table*}

\paragraph{Performance Metrics} Precision, Recall, and F-measure are deployed to evaluate the quality of detected subgraphs by different methods. Higher F-measure reveals better overall performance. For synthetic data, we use the averaged precision, recall, and f-measure over 50 simulated examples.
% For the dataset used in this and following experiments, as the true subgraphs are provided, f-measure that combines precision and recall can be deployed to evaluate the quality of detected subgraphs by different methods.

\paragraph{Comparison Methods and Results} We compare our algorithm with two state of the art baseline methods:  \textit{Meden} \cite{bogdanov2011mining} and \textit{NetSpot} \cite{mongiovi2013netspot}, which were designed specifically for detecting significant anomalous region in dynamic networks and provide implementations. The comparison of results are reported in Table \ref{tab:noise} and Table \ref{tab:evo}. As you can see, our method outperforms these two baseline methods on both synthetic data and real-world data. Both of baselines are heuristic, which can not guarantee the quality of results and cause  worse performance than ours.

\begin{table}[!htbp]
\caption{Results on Washingtong D.C. and Beijing datasets. }\label{tab:evo}
% \vspace{-3pt}
\centering
\setlength\tabcolsep{3pt}
\resizebox{\linewidth}{!}{
\begin{tabular}{|c|c|c|c|c|c|c|}
\hline
\multirow{2}{*}{Methods} & \multicolumn{3}{c|}{Washington D.C.} & \multicolumn{3}{c|}{Beijing} \\ \cline{2-7}
 & \multicolumn{1}{c|}{Precision} & \multicolumn{1}{c|}{Recall} & F-measure & \multicolumn{1}{c|}{Precision} & \multicolumn{1}{c|}{Recall} & F-measure \\ \hline
Meden & 0.7076 & 0.7662 & 0.7342 & 0.6424 & 0.7509 & 0.6882 \\
NetSpot & 0.5823 & 0.7098 & 0.6367 & 0.6789 & 0.7351 & 0.6973 \\
GBGP & 0.7049 & 0.9192 & \textbf{0.7853} & 0.6627 & 0.9634 & \textbf{0.7788} \\ \hline
\end{tabular}
}
\end{table}

\begin{figure*}
    \centering
    \begin{subfigure}[t]{0.24\textwidth}
        \centering
        \includegraphics[width=\linewidth,height=0.7\linewidth]{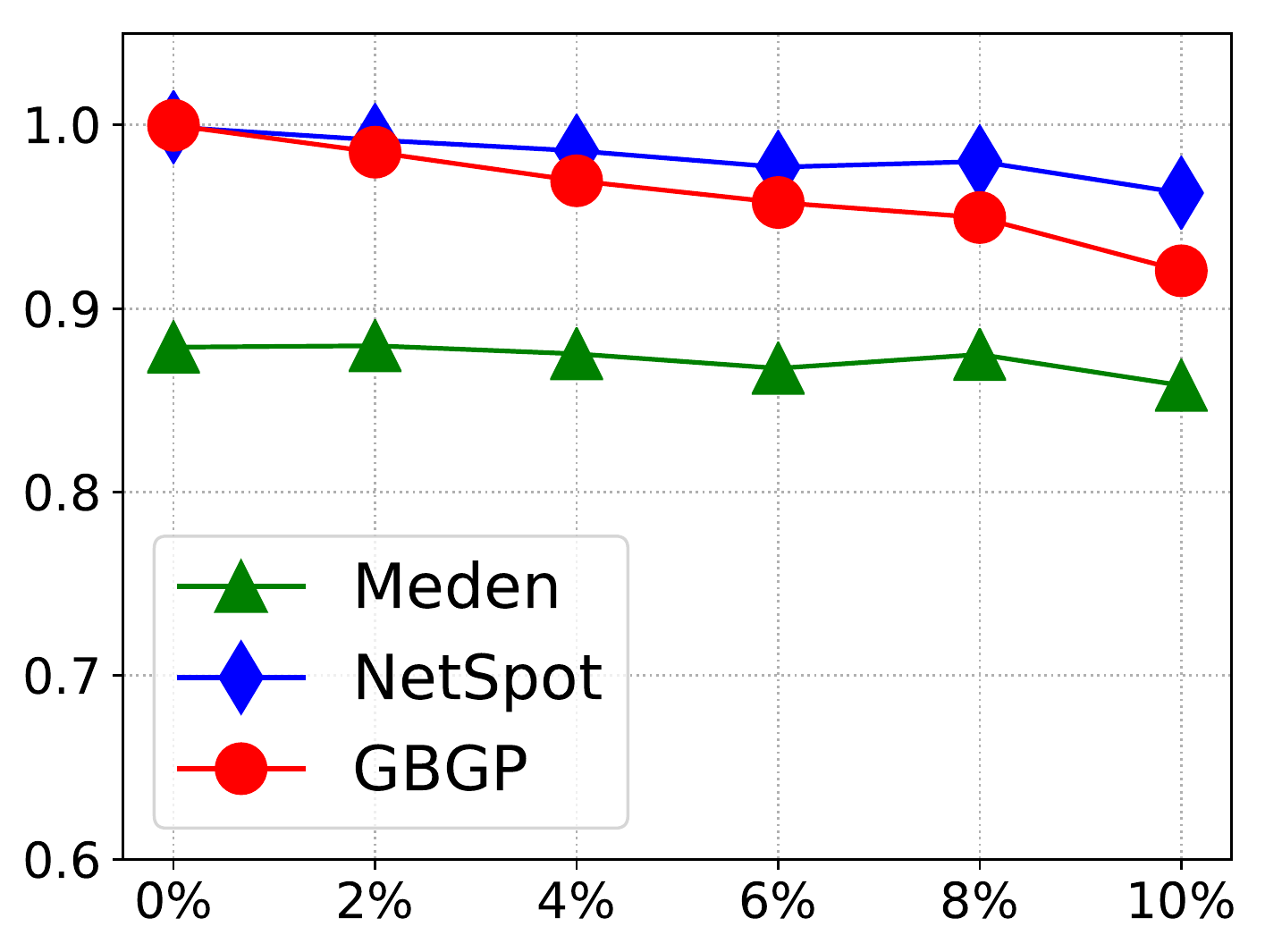}
        \caption{Precision vs Noise Level}
        \label{fig:water:a}
    \end{subfigure}
    \hspace{0pt}
    \begin{subfigure}[t]{0.24\textwidth}
        \centering
        \includegraphics[width=\linewidth,height=0.7\linewidth]{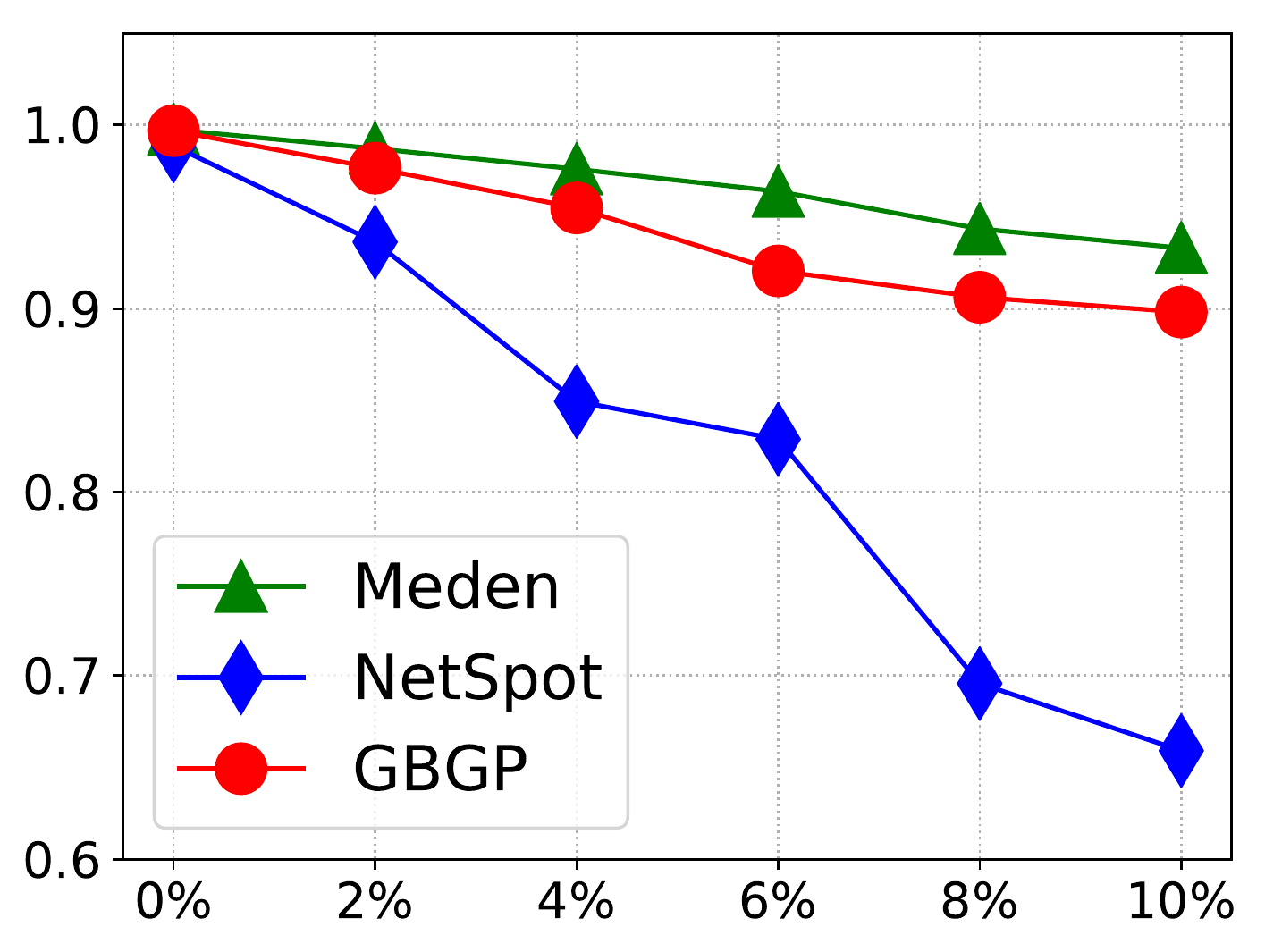}
        \caption{Recall vs Noise Level}
        \label{fig:water:b}
    \end{subfigure}
    \hspace{0pt}
    \begin{subfigure}[t]{0.24\textwidth}
        \centering
        \includegraphics[width=\linewidth,height=0.7\linewidth]{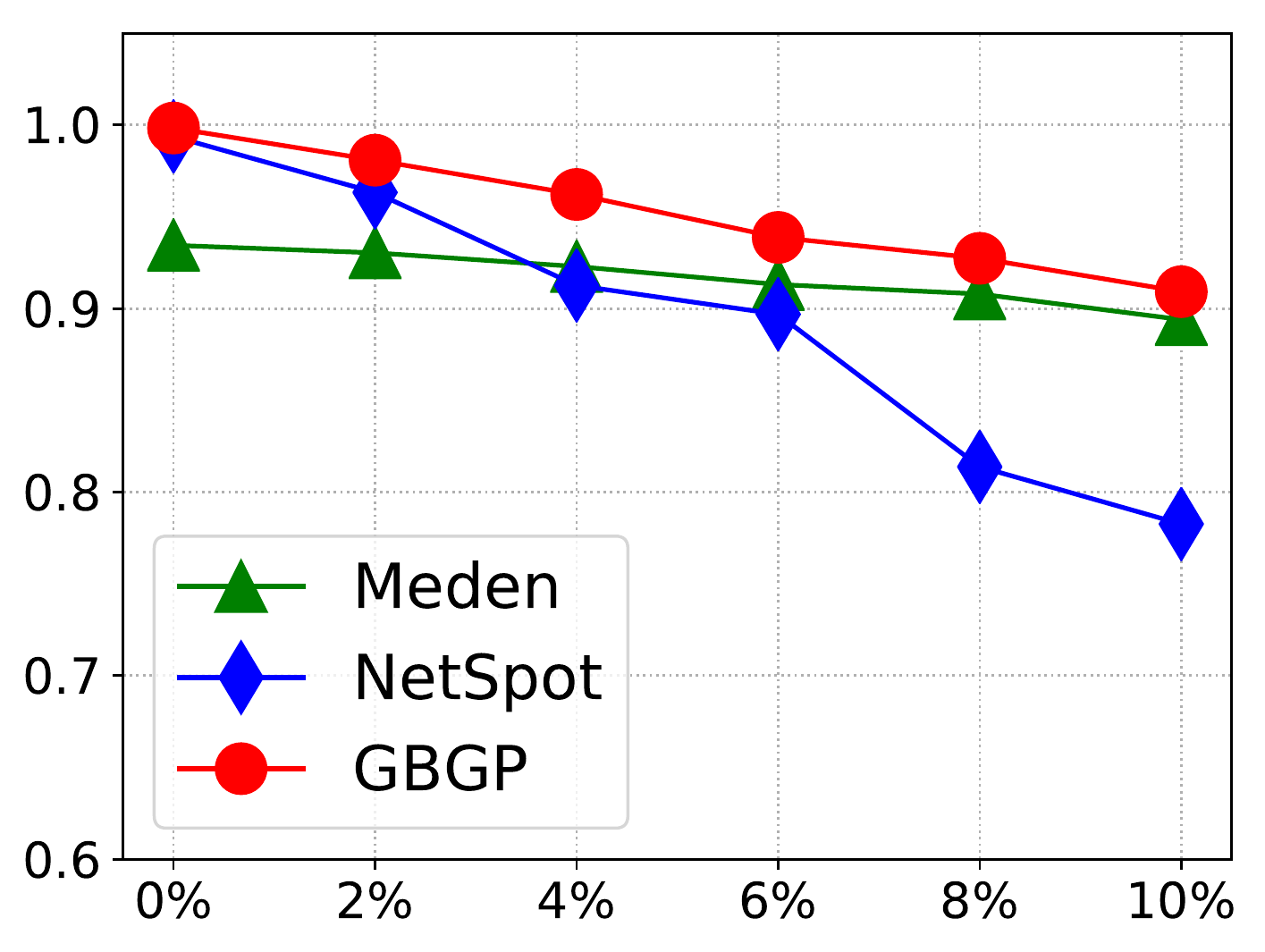}
        \caption{F-measure vs Noise Level}
        \label{fig:water:c}
    \end{subfigure}
    \caption{Precision, Recall, and F-measure curves on Water Pollution dataset with respect to different noise ratios.}
    \label{fig:water:noise}
\end{figure*}

\begin{figure}[ht]
    \vspace{-7pt}
    \centering
    \begin{subfigure}[t]{0.23\textwidth}
        \centering
        \includegraphics[width=\linewidth,height=0.75\linewidth]{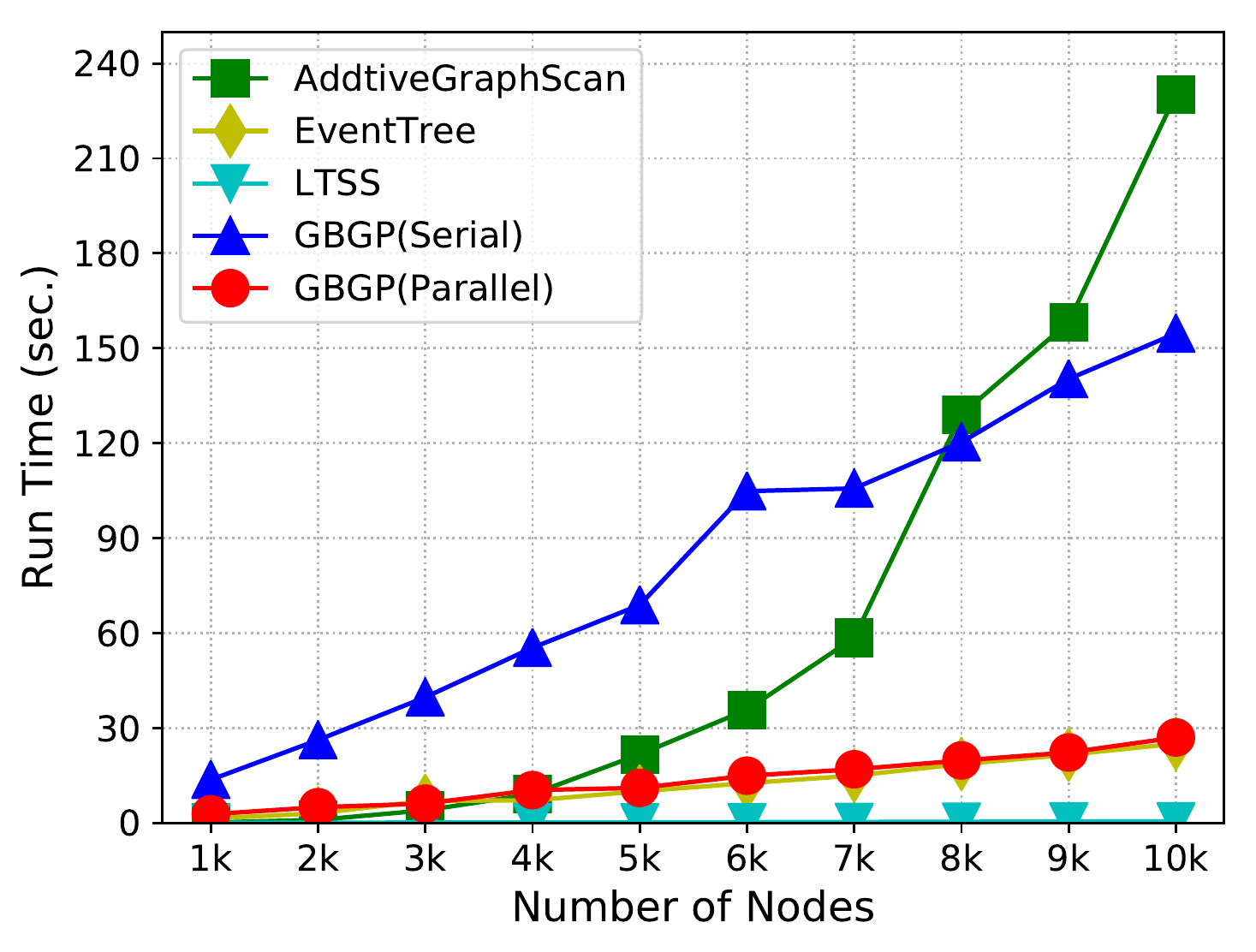}
        \caption{Run Time vs Nodes}
        \label{fig:synthetic:a}
        % \vspace{-5pt}
    \end{subfigure}
    % \hspace{10pt}
    \begin{subfigure}[t]{0.23\textwidth}
        \centering
        \includegraphics[width=\linewidth,height=0.75\linewidth]{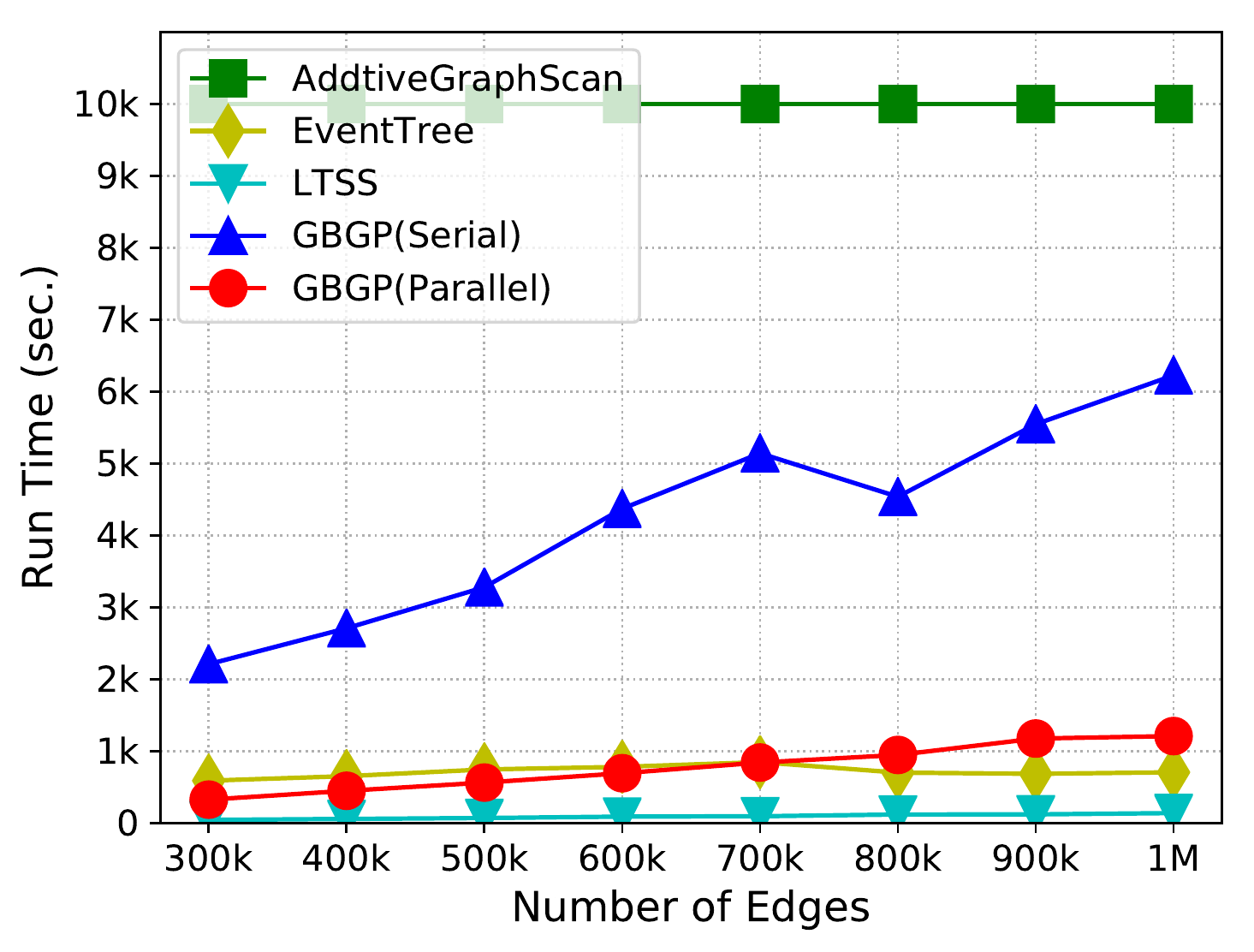}
        \caption{Run Time vs Edges}
        \label{fig:synthetic:b}
        % \vspace{-5pt}
    \end{subfigure}
    \vspace{-3pt}
    \caption{Comparison of run time on synthetic datasets. Figure (a) shows our methods run in nearly linear time w.r.t to the network size, where $|\mathbb{E}|=3|\mathbb{V}|$. Figure (b) shows that our algorithm can be easily scaled up to $1,000,000$ edges with node size $|\mathbb{V}| = 100,000$, by contrast, the AdditiveGraphScan runs over $10,000$ seconds on all cases.}
    \label{fig:synthetic}
    % \vspace{-20pt}
\end{figure}

\paragraph{Robustness Validation} Except for measuring the accuracy of subgraph detection, we also test the robustness of subgraph detection method on water pollution dataset as \cite{shao2017efficient,zhou2016graph}. $P$ percent of nodes are selected randomly, and their sensor binary values are flipped in order to test the robustness of methods to noises, where $P \in \{2, 4, 6, 8, 10\}$. Figure \ref{fig:water:noise} shows the precision, recall, and f-measure of all the comparison methods on the detection of polluted nodes in the water pollution dataset with respect to different noise ratios. The results indicate that our proposed method GBGP is the best overall performance for all of the settings, which verifies the robustness of our method.

% \textcolor{red}{\paragraph{Parameter Tuning} We strictly follow strategies recommended by authors in their original papers to tune the related model parameters. For NetSpot \cite{mongiovi2013netspot}, edges are weighted by comparing their p-value to a significance level threshold $\mu$ (0.01 recommended by authors). Our proposed methods have two parameters, 1) sparseness parameter $s$ (an upper bound of the subgraph size on each block) and 2) trade off parameter $\lambda$. Part of data is extracted from datasets as training dataset and grid search is done to decide those two parameters. $\lambda$ is selected in range $[0.001, 0.1]$ with step size $0.001$. More details are provided in the supplementary material.}
% We select the best $\lambda$ in the range $\{0.0005, 0.001, 0.005, 0.01, 0.05, 0.1\}$, while the range of $s$ depends on specific datasets.

\subsection{Subgraph Dectection in Network of Networks}

\paragraph{Synthetic Datasets} We generate several networks with different network sizes using Barab\'asi-Albert model, and then apply random walk algorithm to simulate the ground-truth subgraph with size as $10\%$ of network size. The nodes in true subgraph have features following normal distribution $\mathbb{N}(5,1)$, and the features of background nodes follows distribution $\mathbb{N}(0,1)$. The synthetic datasets are used for scalability analysis in terms of size of nodes and size of edges, which we denote them as \textbf{\textit{SynNode}} and \textbf{\textit{SynEdge}} respectively.   

\paragraph{Real-world Datasets}
1) \textit{\textbf{Beijing Road Traffic Dataset}}: we use static network data per time stamp from 5PM. to 7PM. in previous application.
2) \textit{\textbf{Wikivote Dataset}}\footnote{\url{https://snap.stanford.edu/data/}\label{data:snap}}: the network contains all the Wikipedia voting data from the inception of Wikipedia till January 2008.
% with $7,115$ nodes and $103,689$ edges. 
3) \textit{\textbf{CondMat Dataset}\textsuperscript{\ref{data:snap}}}: the collaboration network is from the e-print arXiv and covers scientific collaborations between authors papers submitted to Condense Matter category. 
% with $23,133$ nodes and $93,497$ edges.  
For Wikivote and CondMat datasets, we simulate the true subgraphs of size $1,000$ using random walk, and the node attribute in true subgraphs follows distribution $\mathbb{N}(5, 1)$, otherwise $\mathbb{N}(0,1)$. 4) \textit{\textbf{DBLP}}\footnote{\url{http://konect.uni-koblenz.de/networks/dblp_coauthor}}: the collaboration graph of authors of scientific papers from DBLP computer science bibliography. An edge between two authors represents a common publication, and node attribute is the number of publications. We extract a subset of the dataset ranging from year 1995 to 2005. 
% The dataset contains 329,404 nodes and 1,082,106 edges. 
We apply random walk to get subgraphs with size 20,000 and inject the anomalies as our true subgraph as suggested by \cite{mongiovi2013netspot}.  Statistics of all datasets are provided in Table \ref{table:app2}.

\begin{table}[!htbp]
\vspace{-5pt}
\caption{Statistics of Datasets for the 2nd Application.}
\begin{adjustbox}{center}
\resizebox{0.85\linewidth}{!}{
\centering
\begin{tabular}{|c|c|c|c|c|}
\hline
\multirow{2}{*}{Datasets} & \multicolumn{4}{c|}{Statistics} \\ \cline{2-5} 
& \multicolumn{1}{c|}{Node} & \multicolumn{1}{c|}{Edge} & \multicolumn{1}{c|}{Blocks} & \multicolumn{1}{c|}{Processors} \\ \hline
SynNode & 1,000$\sim$10,000 & 3,000$\sim$30,000 & 10 & 10\\
SynEdge & 100,000 & 300,000$\sim$1,000,000 & 100 & 50\\
Beijing & 59,000 & 70,317 & 100 & 50\\
Wikivote & 7,115 & 103,689 & 10 & 10\\
CondMat & 23,133 & 93,497 & 100  & 50\\
DBLP & 329,404 & 1,082,106 & 100 & 50\\ \hline
\end{tabular}\label{table:app2}
}
\end{adjustbox}
\end{table}

\begin{table*}[!htbp]
\caption{Results on Beijing, Wikivote, CondMat and DBLP datasets. The run time is measured in seconds.}\label{tab:block}
% \vspace{-3pt}
\begin{adjustbox}{center}
% \hspace{-15pt}
\centering
\setlength\tabcolsep{2pt}
\resizebox{\textwidth}{!}{
\begin{tabular}{|c|c|c|c|c|c|c|c|c|c|c|c|c|c|c|c|c|}
\hline
\multirow{2}{*}{Method} &
\multicolumn{4}{c|}{Beijing} & 
\multicolumn{4}{c|}{Wikivote} & \multicolumn{4}{c|}{CondMat}& \multicolumn{4}{c|}{DBLP}  \\ \cline{2-17} 
 & \multicolumn{1}{c|}{Precision} & \multicolumn{1}{c|}{Recall} & \multicolumn{1}{c|}{F-measure} & Run Time & \multicolumn{1}{c|}{Precision} & \multicolumn{1}{c|}{Recall} & \multicolumn{1}{c|}{F-measure} & Run Time & \multicolumn{1}{c|}{Precision} & \multicolumn{1}{c|}{Recall} & \multicolumn{1}{c|}{F-measure} & Run Time&
 \multicolumn{1}{c|}{Precision} & \multicolumn{1}{c|}{Recall} & \multicolumn{1}{c|}{F-measure} & Run Time \\ \hline
AddtivegGraphScan &
0.4295 & 0.6884 & 0.5192 & 10846.94 &
0.9543 & 0.9959 & 0.9747 & 249.97 & 0.9753 & 0.9900 & \textbf{0.9826} & 1188.33& / & / & / & / \\
EventTree & 
0.5547 & 0.5577 & 0.5369 & 90.68 & 
0.9088 & 0.9654 & 0.9360 & 80.99 & 0.8623 & 0.9204 & 0.8902 & 100.23& 0.8213 & 0.1922 & 0.3113 & 1961.58 \\
LTSS & 
0.5144 & 0.8333 & 0.6320 & \textbf{7.56} & 
0.9543 & 0.9959 & 0.9747 & \textbf{1.72} & 0.5174 & 1.0000 & 0.6819 & \textbf{3.85}& 0.3910 & 1.0000 & 0.5622 & \textbf{533.13} \\
GBGP(Serial) &
0.9166 & 0.7286 & \textbf{0.8057} & 843.37 &
0.8287 & 0.9908	& 0.90254	& 610.54  &  0.9132	& 0.9859 & 0.9479	& 1243.71 & 0.4701 & 0.9672 & \textbf{0.6354} & 13497.50  \\
GBGP(Parallel) & 
0.9105 & 0.7283 & 0.8028 & 154.12 &
0.9637	& 0.9888 & \textbf{0.9761} & 171.98 & 0.9423 & 0.9835	& 0.9624 &	113.08 & 0.4683 & 0.9672 & 0.6311 & 567.20 \\ \hline
\end{tabular}
}
\end{adjustbox}
\end{table*}
\setlength{\dbltextfloatsep}{10pt}

\paragraph{Performance Metrics} Except for metrics (precision, recall and f-measure) used for evaluating the detection performance, we also compare and report the \textbf{run time} among different methods in this application to evaluate the scalability. 

\paragraph{Comparison Methods and Results} We compare our method with three baselines: 1) \textit{EventTree} \cite{rozenshtein2014event}, 2) \textit{AdditiveGraphScan} \cite{speakman2013dynamic}, and 3) \textit{LTSS} \cite{neill2012fast}, which were designed specifically for event detection on static networks. The average precision, recall, f-measure, as well as run time on all methods are reported in Table \ref{tab:block}. Our method outperforms the baselines in terms of f-measure by the compromise on a small amount of run time. All of baselines have their own shortcomings. Despite AdditiveGraphScan can get comparable performance as our method on some datasets, it is a heuristic algorithm without theoretical guarantees and not scalable for large scale networks. We do not report the result of AdditiveGraphScan on DBLP dataset, since it takes over one day to run and infeasible to tune the parameters. EventTree and LTSS are scalable, but their performances are not as good as our method. 

\paragraph{Scalability Analysis} We evaluate the scalability of different methods in terms of the sizes of nodes and edges. Figure \ref{fig:synthetic} reports the run time of our methods compared with the baseline methods. In order to run our algorithm, we partition the static network into multiple blocks with METIS %\cite{karypis1998software}
\cite{karypis1998fast}, and run the parallel algorithm with multiple processors. Our method is able to get comparable performance as those customized algorithms of this specific problem, and it is more scalable if we properly utilize the computing resource based on network properties.

% \textcolor{red}{\paragraph{Parameters Tuning} For all the parameters used in AddtiveGraphScan, EventTree, and LTSS, we follow the same settings as \cite{zhou2016graph,shao2018efficient,neill2012fast}. For parameters in our method, the same strategy as the previous application is used. We deploy multiple processors to run parallelized GBGP algorithm for different datasets, and more details are provided in Table \ref{table:app2}.
% \textbf{Implementation} We implement all discussed algorithms and perform experiments on 
% 64-bit machines with Intel(R) Xeon(R) CPU E5-2680 v4 @ 2.40GHz and 251GB memory.}

\section{Related Work} \label{section:related}
\textit{a) Subgraph Detection.} Subgraph detection 
% \textcolor{red}{is an important problem in graph mining. Existing} 
methods mainly find subgraphs that satisfy some topological constraints, such as connected subgraphs, dense subgraphs and compact subgraphs, including EventTree \cite{rozenshtein2014event},   NPHGS\cite{chen2014non} for static graphs, Meden \cite{bogdanov2011mining}, NetSpot\cite{mongiovi2013netspot}, and AdditiveGraphScan \cite{speakman2013dynamic}
%, and dGraphScan \cite{shao2018efficient} 
for dynamic graphs, which are all heuristic.
% \textcolor{red}{Those graphs with static network structure are what we focus on.} 
% According to dynamics of attributes on nodes, the relevant studies can be divided into two parts. For static graphs, the most recent methods include EventTree \cite{rozenshtein2014event},   NPHGS\cite{chen2014non}, and method introduced in \cite{qian2014connected}, in which the first two methods are heuristic and the last one provides performance bounds but is not scalable. For dynamic graphs, the latest methods include Meden \cite{bogdanov2011mining}, NetSpot\cite{mongiovi2013netspot}, AdditiveGraphScan \cite{speakman2013dynamic}, and dGraphScan \cite{shao2018efficient}, which are all heuristic.
% \textit{b) Interdependent Networks.} 
% % Complex systems can be modeled as complex networks, which usually encompass many subsystems that interact with or depend on each other. These networks composed of multiple interdependent networks are also known as multilayer networks, networks of networks \cite{kivela2014multilayer}, or multiplex networks \cite{mucha2010community,gomez2012evolution}. 
% % Research in this field mostly studies static and evolving statistical characteristics of structures in these networks, which does not consider attributes on nodes and edges. 
% Typical applications and problems for interdependent networks include: community structure analysis \cite{gao2014single}, the spread of epidemics, and the organizational laws of social interactions \cite{gao2012networks}, among others.
\textit{b) Structured Sparse Optimization.} %Sparsity is effective to model latent structure in high-dimensional data and remains a mathematically tractable concept. 
% Beyond the ordinary sparsity model, a variety of structured sparsity models have been proposed to solve specific structured constraints on signals, such as sparsity models defined through trees, groups, clusters, paths and connected subgraphs \cite{hegde2015nearly,zhou2016graph}.
% \textcolor{red}{Those methods for structured sparsity that focus on general nonlinear cost functions fall into two major categories, including: 1) structured sparsity-inducing norms based and 2) model-projection based, both of which often assume that the cost function $F({\bf x})$ satisfies a certain convexity/smoothness condition, such as Restricted Strong Convexity/Smoothness (RSC/RSS) or Stable Mode-Restricted Hessian (SMRH) \cite{hegde2015approximation, zhou2016graph}.} 
The seminal work on general approximate graph-structured sparsity model is \cite{hegde2015nearly}. 
% The authors present GRAPH-COSAMP, a variant of COSAMP \cite{needell2009cosamp}, for compressive sensing and linear regression problems based on head and tail approximations of $\mathbb{M}(\mathbb{G},s)$. 
General structured optimization methods on single graph was proposed to do subgraph \cite{chen2016generalized,zhou2016graph} or subspace \cite{chen2017generic} detection. 
% We extend the method \cite{chen2016generalized} to handle interdependent networks, which are block-structured. 
% The work \cite{chen2016generalized} can be viewed as a special case of our work, which handles networks with only one block.

\section{Conclusion and Future Work} \label{section:conclusion}
This paper presents a general framework, GBGP, to solve a nonconvex optimization problem subject to graph block-structured constraints in nearly linear time with a theoretical approximation guarantee.
% \textcolor{red}{and it can be applied to a broad class of subgraph detection problems in interdependent networks. A parallel version of our algorithm is also be proposed to improve the scalability.} 
We evaluate our model on two applications, and results of both experiments show that the algorithm enjoys better effectiveness and efficiency than state of the art methods while our work is a general framework and can be used in more scenarios. For future work, we will extend the work on network data with high-dimensional node attributes and different graph topological constraints.

\section*{Acknowledgments}

The first author has been supported by the CSC scholarship. This work has been supported by the National Key Research and Development Program of China under grant 2016YFB1000901, the National Natural Science Foundation of China under grant 91746209, the Program for Changjiang Scholars and Innovative Research Team in University (PCSIRT) of the Ministry of Education of China under grant IRT17R32, and the US National Science Foundation under grant IIS-1815696 and IIS-1750911.

\bibliographystyle{IEEEtran}
\bibliography{IEEEabrv,references}

\end{document}

\maketitle

\begin{abstract}
    This document is a supplementary matrial for the submission of ICDM 2019. We provide more details about the \textbf{proofs}, descriptions and settings about \textbf{experiments} done in the paper, and \textbf{implementation} information.
\end{abstract}

\section{Proofs}

\subsection{Theorem 1}

\begin{definition}[Weak Restricted Strong Convexity Property (WRSC) for Multiple Blocks]\label{def:wrsc}
A function $F({\bf x})$ has condition $(\xi,\delta,\mathbb{M})$-WRSC, if $\forall {\bf x}, {\bf y}\in\mathbb{R}^N$ and $\forall S\in\mathbb{M}$ with $\mathrm{supp}({\bf x})\cup\mathrm{supp}({\bf y})\subseteq S$, the following inequality holds for some $\xi >0$ and $0<\delta<1$:
\begin{equation}
    \|{\bf x}-{\bf y}-\xi\nabla_S F({\bf x})+\xi\nabla_S F({\bf y})\|_2\leq\delta\|{\bf x}-{\bf y}\|_2
\end{equation}
where ${\bf x}=({\bf x}^1,\dots,{\bf x}^K), {\bf y}=({\bf y}^1,\dots,{\bf y}^K), {\bf x}^k, {\bf y}^k\in\mathbb{R}^{N_k}, k=1,\dots,K$,  topological constraint $\mathbb{M}$ can be expressed as $\mathbb{M}(\mathbb{G}, s)=\bigcup_{k=1}^K\mathbb{M}_k(\mathbb{G},s_k), s=\sum_{k=1}^K s_k$, and the subgraph in $k^{\text{th}}$ block is $S_k$, which satisfies $|S_k|\leq s_k, S_k\subseteq \mathbb{G}, S=\bigcup_{k=1}^K S_k, |S|\leq s$.
\end{definition}

\begin{theorem}\label{thm:convege}
Consider the graph block-structured constraint with $K$ blocks $\mathbb{M}(\mathbb{G}, s)=\bigcup_{k=1}^K\mathbb{M}_k(\mathbb{G},s_k)$ and a cost function $F:\mathbb{R}^N\to\mathbb{R}$ that satisfies condition $(\xi,\delta,\mathbb{M}(\mathbb{G},8s))$-WRSC. If $\eta=c_H(1-\delta)-\delta>0$, then for any true ${\bf x}\in\mathbb{R}^N$ with $\mathrm{supp}({\bf x})\in\mathbb{M}((\mathbb{G},s)$, the iteration of Algorithm obeys
\begin{equation}
    \|{\bf x}^{i+1}-{\bf x}^*\|_2\leq\alpha\|{\bf x}^i-{\bf x}^*\|_2+\beta\|\nabla_I F({\bf x}^*)\|_2
\end{equation}
where $c_H=\min_{k=1,\dots,K}\{{c_{H_k}}\}$, $c_T=\max_{k=1,\dots,K}\{{c_{T_k}}\}$, $I=\argmax_{S\in\mathbb{M}}\|\nabla_S F({\bf x}^*)\|_2$, $\alpha=\frac{1+c_T}{1-\delta}\sqrt{1-\eta^2}$, and $\beta=\frac{\xi(1+c_T)}{1-\delta}\left[\frac{1+c_H}{\eta}+\frac{\eta(1+c_H)}{\sqrt{1-\eta^2}}+1 \right]$. $c_{H_k}$ and $c_{T_k}$ denote head and tail projection approximation factors on $k^{\text{th}}$ block.
\end{theorem}

\begin{proof}
Let $\rvr^{i+1}=\rvx^{i+1}-\rvx^*$, $\rvx^i=(\rvx^{1,i},\dots,\rvx^{K,i}), \rvb^i=(\rvb^{1,i},\dots,\rvb^{K,i})$, $\rvx^{k,i},\rvb^{k,i}\in\R^{N_k}$ are coefficient variables for partitioned block $k$ at $i^{\text{th}}$ iteration, $k=1,\dots, K$, ${\bf x}^{*}$ is optimum of ${\bf x}$, $\rvx^i$ refers to the whole coefficient variable at iteration $i$. For block $k$, its head and tail projections are $\Gamma_{\rvx^k}^{i}=H(\nabla_{\rvx^k}F(\rvx^{i}))$ and $\Psi_{\rvx^k}^{i}=T(\rvb^i)$. $\|\rvr^{i+1}\|_2$ is upper bounded as
% \footnote{Here, the superscript in $\rvx^k$ refers to the block of network $k$, $\rvx^i$ refers to the whole coefficient variable when iteration $i$, while $\rvx^*$ refers to ground truth of $\rvx$.}
\begin{align*}
&\|\rvr^{i+1}\|_2\\
=&\|\rvx^{i+1}-\rvx^*\|_2\\
=&\|\rvx^{i+1}-\rvb+\rvb-\rvx^*\|_2\\
\leq&\|\rvx^{i+1}-\rvb\|_2+\|\rvx^*-\rvb\|_2\quad \Leftarrow\sqrt{a+b}\leq\|a\|_2+\|b\|_2\\ 
=&\sqrt{\|\rvx^{1,i+1}-\rvb^{1}\|_2^2+\cdots+\|\rvx^{K,i+1}-\rvb^{K}\|_2^2}+\|\rvx^*-\rvb\|_2\\
\leq&\sqrt{c_{T_1}^2\|\rvx^{1,*}-\rvb^1\|_2^2+\cdots+c_{T_K}^2\|\rvx^{K,*}-\rvb^K\|_2^2}+\|\rvx^*-\rvb\|_2\\
&{\color{red}{(\rvx^k)^* \text{ refers to the optimum of } \|{\bf x}^k-{\bf b}^k\ |_2}}\\
& \text{Tail projection}, \|\rvx-\rvx_S\|\leq c_T\min_{S'\in\sM(\gG, s)}\|\rvx-\rvx_{S'}\|_2\Rightarrow \|\rvx^{k,i+1}-\rvb^k\|_2\leq c_T\|\rvx^{k,*}-\rvb^k\|_2\\
\leq&\sqrt{c_{T_1}^2\|(\rvx^*)^{1}-\rvb^1\|_2^2+\cdots+c_{T_K}^2\|(\rvx^*)^{K}-\rvb^K\|_2^2}+\|\rvx^*-\rvb\|_2\\
&{\color{red}{x^{k,*} \text{ refers to the } k^{\text{th}} \text{ block of global optimum } {\bf x}^* }} \\
=&c_T\sqrt{\|\rvx^{1,*}-\rvb^1\|_2^2+\cdots+\|\rvx^{K,*}-\rvb^K\|_2^2}+\|\rvx^*-\rvb\|_2\quad\text{where } {\color{red}{c_T=\max_{k=1,\dots,K}\{c_{T_k}\}}}\\
=&c_T\|\rvx^*-\rvb\|_2+\|\rvx^*-\rvb\|_2\\
=&(1+c_T)\|\rvx^*-\rvb\|_2\numberthis\label{eq:ri+1}
\end{align*}
where the second inequality follows from the fact that global optimnum of $k^{\text{th}}$ block $({\bf x}^*)^k$ is not necessarily the optimum of $\|{\bf x}^k-{\bf b}^k\|_2$.

Suppose $\Omega=\bigcup_{k=1}^K\Omega_{\rvx^k}$, where $\Omega_{{\bf x}^k}$ is defined in \textbf{Line 4} in Algorithm 1. The component $\|(\rvx^*-\rvb)_{\Omega}\|_2^2$ is upper bounded as
\begin{align*}
&\|(\rvx^*-\rvb)_{\Omega}\|_2^2\\
=&\left\langle \rvb-\rvx^*,(\rvb-\rvx^*)_{\Omega} \right\rangle\\
=&\left\langle \rvb-\rvx^*-\xi\nabla_{\Omega}F(\rvb)+\xi\nabla_{\Omega}F(\rvx^*),(\rvb-\rvx^*)_{\Omega} \right\rangle-\left\langle \xi\nabla_{\Omega}F(\rvx^*),(\rvb-\rvx^*)_{\Omega} \right\rangle\\
&\rvb=\argmin_{\rvx\in\R^N} F(\rvx)\quad \mathrm{s.t.}\quad \mathrm{supp}(\rvx)\subseteq\Omega\Rightarrow{\color{red}{\nabla_{\Omega}F(\rvb)=0}}\\
\leq&\|\rvb-\rvx^*-\xi\nabla_{\Omega}F(\rvb)+\xi\nabla_{\Omega}F(\rvx^*)\|_2\|(\rvb-\rvx^*)_{\Omega}\|_2+\xi\|\nabla_{\Omega}F(\rvx^*)\|_2\|(\rvb-\rvx^*)_{\Omega}\|_2\\
&\text{Cauchy-Schwarz inequality}, |\langle \rvu,\rvv \rangle|\leq\|\rvu\|_2\|\rvv\|_2\\
\leq&\delta\|\rvb-\rvx^*\|_2\|(\rvb-\rvx^*)_{\Omega}\|_2+\xi\|\nabla_{\Omega}F(\rvx^*)\|_2\|(\rvb-\rvx^*)_{\Omega}\|_2\\
&{\color{red}{\text{WRSC property}}}, \|\rvx-\rvy-\xi\nabla_S F(\rvx)+\xi\nabla_S F(\rvy)\|_2\leq \delta\|\rvx-\rvy\|_2
\end{align*}
where the second equality follows from the fact that $\nabla_{\Omega}F(\rvb)=0$ since $\rvb$ is the solution to problem (8) in Algorithm 1, and the last inequality follows from condition $(\xi,\delta,\sM(\sG, 8s)), \sM(\sG, 8s)=\bigcup_{k=1}^K\sM_k(\sG, 8s_k)$. After simplification, we have
\begin{equation*}
\|(\rvx^*-\rvb)_{\Omega}\|_2\leq\delta\|\rvb-\rvx^*\|_2+\xi\|\nabla_{\Omega}F(\rvx^*)\|_2
\end{equation*}
It follows that
\begin{align*}
&\|\rvx^*-\rvb\|_2\leq\|(\rvx^*-\rvb)_{\Omega}\|_2+\|(\rvx^*-\rvb)_{\Omega^c}\|_2\quad \Leftarrow\sqrt{a+b}\leq\|a\|_2+\|b\|_2\\
\leq&\delta\|\rvb-\rvx^*\|_2+\xi\|\nabla_{\Omega}F(\rvx^*)\|_2+\|(\rvx^*-\rvb)_{\Omega^c}\|_2
\end{align*}
After rearrangement, we obtain
\begin{align*}
&\|\rvb-\rvx^*\|_2\\
\leq&\frac{\|(\rvb-\rvx^*)_{\Omega^c}\|_2}{1-\delta}+\frac{\xi\|\nabla_{\Omega}F(\rvx^*)\|_2}{1-\delta}\\
=&\frac{\|\rvx^*_{\Omega^c}\|_2}{1-\delta}+\frac{\xi\|\nabla_{\Omega}F(\rvx)\|_2}{1-\delta}\quad \Leftarrow{\color{red}{\rvb_{\Omega^c}=0}}\Leftarrow\rvb=\argmin_{\rvx\in\R^N}F(\rvx)\quad\mathrm{s.t.}\quad\mathrm{supp}(\rvx)\subseteq\Omega\\
=&\frac{\|(\rvx^*-\rvx^i)_{\Omega^c}\|_2}{1-\delta}+\frac{\xi\|\nabla_{\Omega}F(\rvx^*)\|_2}{1-\delta}\quad\Leftarrow{\color{red}{\rvx^i_{\Omega^c}=0}} \Leftarrow \Gamma\cup \mathrm{supp}(\rvx^i)=\Omega\\
=&\frac{\|\rvr_{\Omega^c}^i\|_2}{1-\delta}+\frac{\xi\|\nabla_{\Omega}F(\rvx^*)\|_2}{1-\delta}\\
\leq&\frac{\|\rvr_{\Gamma^c}^i\|_2}{1-\delta}+\frac{\xi\|\nabla_{\Omega}F(\rvx^*)\|_2}{1-\delta}\quad\Leftarrow \Omega^c\subseteq\Gamma^c\Leftarrow\Gamma\subseteq\Omega\Leftarrow \Gamma\cup\mathrm{supp}(\rvx^i)=\Omega \numberthis\label{eq:ri}
\end{align*}
where the first equality follows from the fact that $\mathrm{supp}(\rvb)\subseteq\Omega$, the second and last inequalities follow from the fact that $\Omega=\Gamma\cup\mathrm{supp}(\rvx^i)$. Combining inequality (\ref{eq:ri+1}) and above inequality (\ref{eq:ri}), we obtain
\begin{align*}
\|\rvr^{i+1}\|_2\leq&(1+c_T)\frac{\|\rvr_{\Omega^c}^i\|_2}{1-\delta}+(1+c_T)\frac{\xi\|\nabla_{\Omega}F(\rvx^*)\|_2}{1-\delta}\\
\leq&(1+c_T)\frac{\|\rvr_{\Gamma^c}^i\|_2}{1-\delta}+(1+c_T)\frac{\xi\|\nabla_{\Omega}F(\rvx^*)\|_2}{1-\delta}
\end{align*}
From Lemma \ref{lem:bound}, we have
\begin{equation*}
\|\rvr_{\Gamma^c}^i\|_2\leq \sqrt{1-\eta^2}\|\rvr^i\|_2+\left[\frac{\xi(1+c_H)}{\eta}+\frac{\xi\eta(1+c_H)}{\sqrt{1-\eta^2}}\right]\|\nabla_{I}F(\rvx^*)\|_2
\end{equation*}
Combining the above inequalities, we prove the theorem.
\end{proof}

\begin{lemma}\label{lem:wrsc}
Assume that $F$ is a differentiable function. If $F$ satisfies condition $(\xi, \delta,\sM)$-WRSC, then $\forall \rvx,\rvy\in\R^N$ with $\mathrm{supp}(\rvx)\cup\mathrm{supp}(\rvy)\subseteq S\in\sM$, the following two inequalities hold \cite{yuan2014gradient}
\begin{gather*}
\frac{1-\delta}{\xi}\|\rvx-\rvy\|_2\leq\|\nabla_S F(\rvx)-\nabla_S F(\rvy)\|_2\leq\frac{1+\delta}{\xi}\|\rvx-\rvy\|_2\\
F(\rvx)\leq F(\rvy)+\langle \nabla F(\rvy), \rvx-\rvy \rangle+\frac{1+\delta}{2\xi}\|\rvx-\rvy\|_2^2
\end{gather*}
\end{lemma}

\begin{lemma}\label{lem:bound}
Let $\rvr^i=\rvx^i-\rvx^*$ and $\Gamma=\bigcup_{k=1}^K\Gamma_{\rvx^k}, \Gamma_{\rvx^k}\in\sM_k(\sG_k, 2s_k)$ $\rvr_{\rvx^k}^i=\rvx^{k,i}-\rvx^{k,*}$, and $\Gamma_{\rvx^k}=H(\nabla_{\rvx^k}F(\rvx^i))$. Then 
\begin{equation*}
\|\rvr_{\Gamma^c}^i\|_2\leq\sqrt{1-\eta^2}\|\rvr^i\|_2+\left[\frac{\xi(1+c_{H})}{\eta}+\frac{\xi\eta(1+c_H)}{\sqrt{1-\eta^2}} \right]\|\nabla_{I}F(\rvx^*)\|_2
\end{equation*}
where $\eta=c_H(1-\delta)-\delta$ and $I=\argmax_{S\in\sM(\sG, 8s)}\|\nabla_S F(\rvx^*)\|_2$. We assume that $c_H$ and $\delta$ are such that $\eta>0$.
\end{lemma}

\begin{proof}
Denote $\Phi=\mathrm{supp}(\rvx^*)\in\sM=\bigcup_{k=1}^K\sM_k(\sG, s_k), \Gamma_{x^k}=H(\nabla_{\rvx^k}F(\rvx)), \Gamma=\bigcup_{k=1}^K, \Gamma_{\rvx^k}\in\sM_k(\sG, 2s_k), \rvr^i=\rvx^i-\rvx^*$, and $\Omega=\mathrm{supp}(\rvr^i)\in\sM(\sG, 6s)=\bigcup_{k=1}^K\sM_k(\sG,6s_k))$. The component $\|\nabla_{\Gamma}F(\rvx^i)\|_2$ can be lower bounded as
\begin{align*}
&\|\nabla_{\Gamma} F(\rvx^i)\|_2\\
=&\sqrt{\left\|\left[\nabla_{\Gamma_{\rvx^1}}F(\rvx^i),\dots,\nabla_{\Gamma_{\rvx^K}}F(\rvx^i)\right]\right\|_2^2}\\
=&\sqrt{\|\nabla_{\Gamma_{\rvx^1}}F(\rvx^i)\|_2^2+\cdots+\|\nabla_{\Gamma_{\rvx^K}}F(\rvx^i)\|_2^2}\\
\geq&\sqrt{c_{H_1}^2\|\nabla_{\Phi^1}F(\rvx^i)\|_2^2+\cdots+c_{H_K}^2\|\nabla_{\Phi^K}F(\rvx^i)\|_2^2}\\
&\text{Head projection}, \|\rvx_S\|_2\geq c_H\max_{S'\in\sM(\sG, s)}\|\rvx_{S'}\|_2\\
\geq&c_H\sqrt{\|\nabla_{\Phi^1}F(\rvx^i)\|_2^2+\cdots+\|\nabla_{\Phi^K}F(\rvx^i)\|_2^2}\quad c_H=\min\{c_{H_k}\}, k=1,\dots,K\\
=&c_H\|\nabla_{\Phi}F(\rvx^i)\|_2\\
=&c_H\|\nabla_{\Phi}F(\rvx^i)-\nabla_{\Phi}F(\rvx^*)+\nabla_{\Phi}F(\rvx^*)\|_2\\
\geq& c_H(\|\nabla_{\Phi}F(\rvx^i)-\nabla_{\Phi}F(\rvx^*)\|_2-\|\nabla_{\Phi}F(\rvx^*)\|_2)\quad\Leftarrow\|a+b\|_2\geq\|a\|_2-\|b\|_2\\
\geq&c_H(1-\delta)\|\rvx^i-\rvx^*\|_2/\xi-c_H\|\nabla_{\Phi} F(\rvx^*)\|_2\quad\Leftarrow \text{Lemma } \ref{lem:wrsc}\\
\geq&c_H(1-\delta)\|\rvr^i\|_2/\xi-c_H\|\nabla_I F(\rvx^*)\|_2 \numberthis\label{bound:geq}
\end{align*}
where the last inequality follows from Lemma \ref{lem:wrsc}. The component $\|\nabla_{\Gamma}F(\rvx^i)\|_2$ can also be upper bounded as
\begin{align*}
\|\nabla_{\Gamma}F(\rvx^i)\|_2\leq&\frac{1}{\xi}\|\xi\nabla_{\Gamma}F(\rvx^i)-\xi\nabla_{\Gamma}F(\rvx^*)\|_2+\|\nabla_{\Gamma}F(\rvx^*)\|_2\quad \Leftarrow \|a+b\|_2\leq\|a\|_2+\|b\|_2\\
\leq&\frac{1}{\xi}\|\xi\nabla_{\Gamma}F(\rvx^i)-\xi\nabla_{\Gamma}F(\rvx^*)-\rvr_{\Gamma}^i+\rvr_{\Gamma}^i\|_2+\|\nabla_{\Gamma}F(\rvx^*)\|_2\\
\leq&\frac{1}{\xi}\left(\|\xi\nabla_{\Gamma\cup\Omega}F(\rvx^i)-\xi\nabla_{\Gamma\cup\Omega}F(\rvx^*)-\rvr_{\Gamma\cup\Omega}^i\|_2+\|\rvr_{\Gamma}^i\|_2\right)+\|\nabla_{\Gamma}F(\rvx^*)\|_2\\
\leq&\frac{\delta}{\xi}\cdot\|\rvr^i\|_2+\frac{1}{\xi}\|\rvr_{\Gamma}^i\|_2+\|\nabla_I F(\rvx^*)\|_2\numberthis\label{bound:leq}\\
&\text{Definition \ref{def:wrsc} } (\xi,\delta,\sM(\sG,s))-\text{WRSC}
\end{align*}
where the last inequality follows from condition $(\xi,\delta,\sM(\sG,s))$-MWRSC and the fact that $\rvr_{\Gamma\cup\Omega}^i=\rvr^i$. Let $\eta=(c_H\cdot(1-\delta)-\delta)$. Combining the two bounds (\ref{bound:geq}) and (\ref{bound:leq}) and grouping terms, we have $\|\rvr_{\Gamma}^i\|\geq\eta\|\rvr^i\|_2-\xi(1+c_H)\|\nabla_I F(\rvx^*)\|_2$. After a number of algebraic manipulations similar to those used in \cite{hegde2014approximation} 
page 11, we prove the lemma.

\end{proof}

\subsection{Theorem 2}

We have proved this theorem in the paper. Here are some details omitted in the paper due to page limit.

Suppose ${\bf x}$ starts from ${\bf 0}$, i.e., ${\bf x}^0={\bf 0}$. Then the $i^{th}$ iteration of Algorithm 1 satisfies
\begin{align*}
    &\|{\bf x}^*-{\bf x}^i\|_2\\
    \leq&\alpha\|{\bf x}^*-{\bf x}^{i-1}\|_2+\beta\|\nabla_I F({\bf x}^*)\|_2\\
    \leq&\alpha^t\|{\bf x}^*-{\bf x}^{i-1}\|_2+(\alpha^{t-1}\beta+\alpha^{t-2}\beta+\cdots+\beta)\|\nabla_I F({\bf x}^*)\|_2\quad \text{after $t$ iterations}\\
    \leq&\cdots\\
    \leq&\alpha^i\|{\bf x}^*-{\bf x}^0\|_2+\frac{\beta}{1-\alpha}\|\nabla_I F({\bf x}^*)\|_2\\
    &\text{the coeffcient of the second term is a geometric sequence and when}\\
    & t\to\infty, \alpha^{t-1}\beta+\alpha^{t-2}\beta+\cdots+\beta=\frac{\beta(1-\beta^{t-1})}{1-\beta}=\frac{\beta}{1-\alpha} \\
    \leq&\alpha^i\|{\bf x}^*\|_2+\frac{\beta}{1-\alpha}\|\nabla_I F({\bf x}^*)\|_2\quad \Leftarrow {\bf x}^0={\bf 0}
\end{align*}

\section{Experiments}

\subsection{Datasets}

\subsection{Synthetic Dataset for Anomalous Evolving Subgraph Detection}
We generate networks using Barab\'asi-Albert preferential attachment model \cite{barabasi1999emergence}. We simulate the evolving true subgraphs in time window $T = 7$, and the true subgraphs in two consecutive time stamps have $50\%$ of overlap. The univariate feature values of background nodes and true nodes are randomly generated in $\mathbb{N}(0,1)$ and $\mathbb{N}(\mu, 1)$ distributions, respectively. We generate multiple temporal networks with different noise level by setting $\mu = [3,4,5]$ for true nodes. As we can see, when $\mu$ is small, the data is more noised, and it is more difficult to distinguish between the anomalous nodes and normal nodes based on the uni-variate feature.  

\subsection{Water Pollution Dataset}
The "Battle of the Water Sensor Networks" (BWSN) \cite{ostfeld2008battle} provides a real world network with 12,527 vertices. Among them, there are 25 vertices with chemical contaminant plumes that are distributed in four different areas. The spread of these contaminant plumes on graph is simulated using the water network simulator EPANET that is used in BWSN for a period of 8 hours. For each hour, each vertex has a sensor that reports 1 if it is polluted; otherwise, reports 0. Except measuring the accuracy of subgraphs detection, we also test the robustness of subgraph detection methods to noises. We randomly selected K percent vertices, and flipped their sensor binary values, where K = 0, 2, 4, ...,10 as \cite{chen2016generalized,shao2017efficient}. The objective is to detect the set of polluted vertices. In order to apply nonparametric graph scan baseline methods to this dataset, we map the sensors whose report values are 1s to the empirical p-value 0.15, and those whose report values are 0s to 1.0. 

\subsection{Washington D.C. Road Traffic Dataset}
We collected traffic data of Washington D.C from June
1, 2013 to March 31, 2014 from INRIX \footnote{http://inrix.com/publicsector.asp.}. The raw INRIX dataset provides traffic speed and reference speed information for each road link per hour
interval. A reference speed is defined as the "uncongested
free flow speed" for each road segment. It is calculated
based upon the 85-th percentile of the measured speed for
all time periods over a few years by INRIX, where the reference speed
serves as a threshold separating two traffic states, congested vs.
uncongested, which will also be served as ground-truth of our experiments. The road traffic dataset for each of the two cities
has 43 weeks in total. An hour is represented by a specific
combination of hour of day $(h \in {6, 7, ..., 22})$, day of week
$(d \in {1, 2, 3, 4, 5})$, and week $(w \in {1, 2, ..., 43}): (h, d, w)$.
We only considered work days from Monday $(d = 1)$ to Friday
$(d = 5)$ and hours from 6AM $(h = 6)$ to 10PM $(h = 22)$. 

\subsection{Beijing Road Traffic Dataset}
Beijing road traffic dataset is comprised of 148,110 nodes and 196,307 edges. The road network covers a $ 40\times 50 $ km spatial range, with a total length (of road segments) of 21, 895 km. Due to missing values in the raw data, we  extract and only use the largest connected component in the network, which contains 59,000 nodes and 70,317 edges. The dataset contains the real-time traffic conditions of four days, from Sep. \nth{12}, 2013 to Sep. \nth{15}, 2013, consisting two workdays and two holidays. In each date, the data consists of traffic records in 144 time-stamps with time resolution as 6 minutes. We only use the data between morning 5AM to evening 10PM, which totally contains $108$ time stamps. The traffic information on each road segment containing the average travel speed $ \bar{\nu} $, the standard deviation of the travel speed $\sigma$, and the number of vehicles, is derived from the GPS trajectories generated by taxicabs traversing the road segment in the time slot. We simply use the difference between reference speed and current speed as edge feature, and obtain the node feature by averaging the edge features associated with that node. 

\subsection{Synthetic Dataset for Subgraph Detection in Network of Networks}
We generate several networks with different network sizes with Barab\'asi-Albert model. We use random walk to simulate the ground-truth subgraph with size as $10\%$ of network size. The nodes in true subgraph have features following normal distribution $\mathbb{N}(5,1)$, and the features of background nodes follows distribution $\mathbb{N}(0,1)$. 

\subsection{Wikivote Dataset}
 \textit{Wikivote} \footnote{\url{https://snap.stanford.edu/data/wiki-Vote.html}} is a network contains all the Wikipedia voting data from the inception of Wikipedia till January 2008, with $7,115$ nodes and $103,689$ edges. We only use the network structure of this data, and we simulate the true subgraphs of size $1000$ using random walk, and the node attribute in true subgraphs follows normal distribution $\mathbb{N}(5, 1)$, otherwise $\mathbb{N}(0,1)$. 

\subsection{CondMat Dataset}
\textit{CondMat}\footnote{\url{https://snap.stanford.edu/data/ca-CondMat.html}} is a collaboration network from the e-print arXiv and covers scientific collaborations between authors papers submitted to Condense Matter category, with $23,133$ nodes and $93,497$ edges. We only use the network structure of this data, and we simulate the true subgraphs of size $1000$ using random walk, and the node attribute in true subgraphs follows normal distribution $\mathbb{N}(5, 1)$, otherwise $\mathbb{N}(0,1)$.

\section{Implementation}

\subsection{}

We implement our algorithm and conduct the experiments in python 2.7. The python modules are required to install before running the code:
\begin{displayquote}
pip install numpy networkx  sparse-learning 
\end{displayquote}

The sparse-learning module provides the implementation of head and tail projection. Our proposed GBMP have two parameters $k$ and $\lambda$, which denote by an upper bound of the subgraph size and trade-off parameter, respectively. We grid search the optimal values of these parameters on the training data, and apply them to the testing data. The provided implementation only outputs the result for one single case, however, our results show in the paper are averaged results over multiple cases.

\bibliographystyle{plain}
\bibliography{references}